\documentclass[11pt]{article}

\usepackage[preprint]{acl}

\usepackage{times}
\usepackage{latexsym}
\usepackage{amsmath}
\usepackage{amsfonts}
\usepackage{enumitem}
\usepackage{rotating}
\usepackage[T1]{fontenc}
\usepackage{booktabs}
\usepackage{multirow}
\usepackage{graphicx}
\usepackage{subcaption}
\usepackage{hyperref}
\usepackage{mdframed}
\usepackage{listings}
\usepackage{tabularx}
\usepackage{amssymb}
\usepackage{algorithm}
\usepackage{algpseudocode}
\usepackage[table]{xcolor}

\definecolor{lightskyblue}{rgb}{0.53, 0.81, 0.98}
\usepackage{tcolorbox}
\definecolor{basegray}{HTML}{F2F2F2}
\definecolor{gainblue}{HTML}{EAF2FF}
\definecolor{lossred}{HTML}{FDECEC}
\definecolor{oursblue}{HTML}{DCEBFF}

\newcommand{\basecell}[1]{\cellcolor{basegray}#1}
\newcommand{\accgain}[2]{\cellcolor{gainblue}#1{\scriptsize\,(+#2)}}
\newcommand{\accloss}[2]{\cellcolor{lossred}#1{\scriptsize\,(-#2)}}
\newcommand{\tokgain}[2]{\cellcolor{gainblue}#1{\scriptsize\,(-#2)}}
\newcommand{\tokloss}[2]{\cellcolor{lossred}#1{\scriptsize\,(+#2)}}
\newcommand{\neutralcell}[2]{\cellcolor{basegray}#1{\scriptsize\,(#2)}}

\newcommand{\oursaccgain}[2]{\cellcolor{oursblue}\textbf{#1}{\scriptsize\,(+#2)}}
\newcommand{\ourstokgain}[2]{\cellcolor{oursblue}\textbf{#1}{\scriptsize\,(-#2)}}

\tcbuselibrary{breakable, skins, listings}

\usepackage[utf8]{inputenc}

\usepackage{microtype}

\usepackage{inconsolata}

\usepackage{graphicx}
\newcommand{\ourmethod}{\textsc{ATLAS}}

\title{{\sc ATLAS:} Verifier-Guided Adaptive Latent \\Activation Steering for Efficient LLM Reasoning}

\author{
  Tuc Nguyen\;\;\; Thai Le \\
  Indiana University \\
  Bloomington, USA\\
  \texttt{\{tucnguye, tle\}@iu.edu}
}

\begin{document}
\maketitle
\begin{abstract}
Recent work on activation and latent steering has demonstrated that modifying internal representations can effectively guide large language models (LLMs) toward improved reasoning and efficiency without updating model parameters. However, most existing approaches rely on fixed steering policies and static intervention strengths, which limit their robustness across problem instances and often result in over- or under-steering. We propose \textsc{Adaptive Test-time Latent Steering} ({\ourmethod}), a lightweight framework that dynamically controls steering decisions at inference time using \textit{a trained, lightweight verifier over the latent states}. Given intermediate hidden states, the verifier predicts the quality of ongoing reasoning and adaptively selects which steering action to apply, enabling per-example and per-step adjustment with minimal overhead. {\ourmethod} provides a unified framework for combining learned latent verification with test-time activation steering, enabling adaptive reasoning control without additional LLM decoding or inference-time process reward model calls. Experiments on multiple mathematical and coding reasoning benchmarks show that {\ourmethod} consistently outperforms both vanilla decoding and fixed steering baselines, achieving higher accuracy while substantially reducing test-time token usage. These results demonstrate that verifier-guided latent adaptation provides an effective and scalable mechanism for controlling reasoning efficiency without sacrificing solution quality. All source code will be publicly available.
\end{abstract}

\section{Introduction}
\begin{figure*}[tb!]
    \centering
    \includegraphics[width=\textwidth]{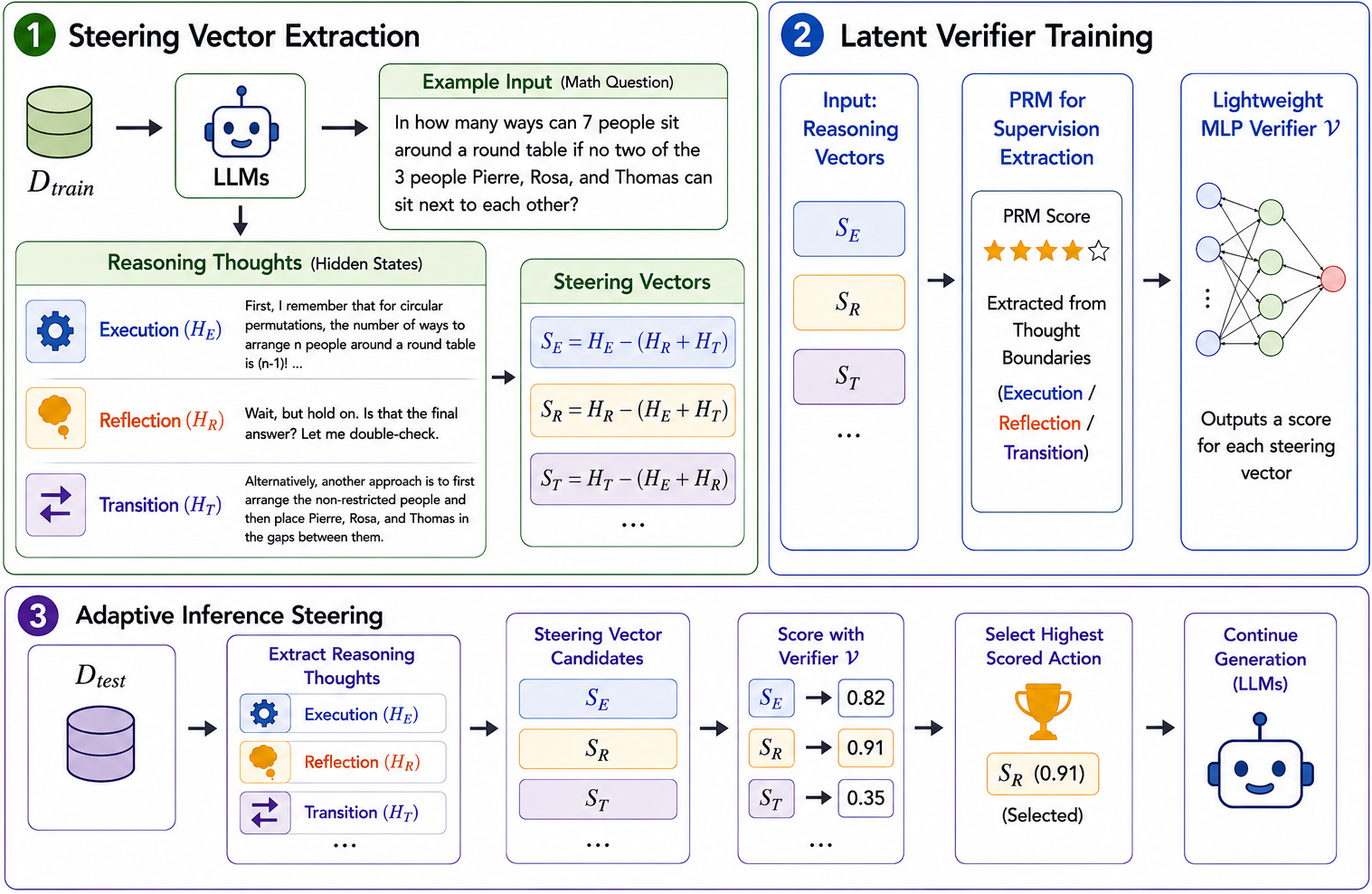}
    \caption{Overview of {\ourmethod}. Offline, {\ourmethod} extracts contrastive action vectors for execution, reflection, and transition by comparing each reasoning mode against its complement, and trains a lightweight latent verifier from PRM-supervised hidden states. During inference, {\ourmethod} scores candidate latent interventions at thought boundaries and applies the action with the highest predicted quality.}
    \label{fig:atlas_overview}
    \vspace{-10pt}
\end{figure*}

Large language models (LLMs) achieve strong performance on complex reasoning tasks such as mathematical problem solving, planning, and code generation~\cite{valmeekam2023planning, alphaproof2024ai, ahn2024large, chen2025advancing, 10.1145/3715754}. Much of this progress relies on eliciting explicit intermediate reasoning, for example, through chain-of-thought prompting and related structured reasoning methods~\cite{wei2022chain, yao2023tree, besta2024graph}. However, longer reasoning is not always better. Recent work shows that LLMs can generate unnecessarily long, redundant, or repetitive reasoning traces, sometimes continuing after a useful solution has already been reached~\cite{fu2024efficiently, chen2024not, wang2025thoughts}. Such behavior increases inference cost and can even degrade final-answer accuracy when the model drifts into unproductive verification loops or unnecessary detours.

Activation and latent steering provide a promising way to control reasoning behavior without updating model parameters via finetuning \cite{houlsby2019parameter, nguyen2024generalizability, nguyen-le-2024-adapters}. By modifying internal representations along directions associated with specific behaviors, steering methods can encourage more concise or effective reasoning at test time~\cite{turner2023steering, chen2025seal, azizi2025activation, zhang2025understanding,nguyen2026beyond}. However, most existing approaches apply a fixed steering direction or intervention strength throughout generation. This static design is poorly matched to multi-step reasoning, where \textit{different stages may require different strategies}: direct execution for straightforward computation, reflection for checking uncertain steps, and transition for escaping unproductive trajectories. Prior analyses further suggest that LLM reasoning traces contain functionally distinct thought units, including execution, reflection, and transition behaviors~\cite{do2025defines,chen2025seal}. When a single steering mode is applied uniformly, these behaviors can become misaligned with the local needs of the problem, causing the model to under-steer difficult examples, over-steer easy examples, or enter repetitive reasoning patterns. As an illustrative example, Appendix~\ref{repetitive_reasoning} shows a case where fixed execution steering causes the model to repeatedly restate the same intermediate computation without progressing to a final answer. This failure motivates adaptive steering: the model should assess its current latent reasoning state and select an appropriate intervention at each reasoning boundary.

We argue that effective reasoning control should be adaptive at the level of individual reasoning steps. Intermediate hidden states of LLMs encode rich semantic, task-specific, and uncertainty-related information, suggesting that they can provide early signals about the quality of an ongoing reasoning trajectory. Instead of generating multiple textual continuations and scoring them with an external process reward model (PRM) directly on the output texts, we ask whether a lightweight verifier can predict reasoning quality \textit{directly from hidden states} and use this signal to select steering actions before generation continues. If this can be achieved, we can significantly reduce runtime by leveraging the comparatively lightweight processing of vectors instead of processing lengthy reasoning texts.

To this end, we propose \textsc{Adaptive Test-time Latent Steering} ({\ourmethod}), a verifier-guided framework for adaptive reasoning control. Offline, ATLAS segments model-generated reasoning traces into thought units, extracts hidden states at thought boundaries, constructs contrastive steering vectors for execution, reflection, and transition modes, and distills PRM-provided step-quality scores into a compact latent verifier. Online, ATLAS evaluates a small action set consisting of no intervention, execution, reflection, and transition. At each thought boundary, the verifier scores the candidate hidden states induced by these actions, and the model continues generation under the highest-scoring intervention. Figure~\ref{fig:atlas_overview} illustrates this offline-online workflow. This allows ATLAS to adapt steering decisions across examples and reasoning steps while avoiding additional LLM decoding or inference-time PRM calls. We evaluate ATLAS on mathematical reasoning and code-generation benchmarks across multiple reasoning models. Compared with vanilla decoding and fixed steering baselines, ATLAS improves final-answer accuracy while substantially reducing generated tokens. We further show that latent verification captures much of the benefit of text-level PRM verification at a lower inference cost, and that adaptive action selection outperforms fixed execution-only steering. Our main contributions are:

\begin{enumerate}[leftmargin=\dimexpr\parindent-0\labelwidth\relax,noitemsep,topsep=0pt]
\item \textbf{Latent process verification.} We distill PRM-provided step-quality supervision into a lightweight verifier that predicts reasoning quality directly from hidden states at thought boundaries.
\item \textbf{Adaptive latent steering.} We formulate test-time steering as action selection over \textit{no intervention}, \textit{execution}, \textit{reflection}, and \textit{transition} reasoning strategy, enabling per-step reasoning control without additional LLM decoding or inference-time PRM calls.
\item \textbf{Empirical analysis across models and tasks.} We evaluate {\ourmethod} across in-domain and cross-domain reasoning benchmarks, showing improved accuracy--efficiency trade-offs and analyzing the role of action selection, verifier reliability, and intervention choices.
\end{enumerate}

\section{Related Work}
{\ourmethod} builds on work in explicit reasoning, activation steering, process supervision, and adaptive inference-time control.
\paragraph{Explicit and Latent Reasoning in LLMs.}
Step-by-step reasoning methods, including Chain-of-Thought (CoT), Tree-of-Thought (ToT), and Graph-of-Thought (GoT), have become standard techniques for eliciting structured reasoning in LLMs~\cite{wei2022chain, yao2023tree, besta2024graph}. Extensions such as self-consistency and least-to-most prompting further improve performance by aggregating multiple reasoning paths or decomposing complex problems into simpler subproblems~\cite{yoon2025reasoning, fu2025deep, zhou2022least}. However, these methods operate primarily through explicit text generation and often require additional generated paths and decomposition steps, making them costly when used for adaptive control. This limits their ability to adapt to local changes in problem-solving difficulty. Recent work on latent reasoning instead studies how reasoning can be represented or manipulated directly in hidden states, offering a more efficient alternative to fully explicit token-level deliberation~\cite{zhu2025surveylatentreasoning, chen2025reasoning, hao2024training}. Our work builds on this perspective by using hidden states not only as reasoning representations, but also as feedback signals for adaptive test-time control.

\paragraph{Activation Steering and Representation Editing.}
Activation steering controls model behavior by modifying internal representations, often through steering vectors derived from contrastive activation differences~\cite{turner2023steering}. Such methods have been used to influence attributes including truthfulness, safety, style, and reasoning behavior without updating model parameters~\cite{burns2022discovering, stolfo2024improving}. Recent reasoning-oriented methods apply similar interventions to reduce inefficient reasoning patterns or encourage specific cognitive modes~\cite{chen2025seal, azizi2025activation, zhang2025understanding}. Despite their effectiveness, most existing approaches apply fixed steering directions or fixed intervention strengths throughout the generation. This static design is poorly suited to multi-step reasoning, where different stages may require execution, reflection, or strategic transition. In contrast, {\ourmethod} dynamically selects among multiple steering actions based on the current latent reasoning state.

\paragraph{Process Supervision and Verifier-Guided Inference.}
Process reward models (PRMs) provide step-level supervision for evaluating intermediate reasoning, offering finer-grained feedback than outcome-only rewards~\cite{lightman2023let, wang2024math}. PRM quality has been shown to correlate strongly with downstream reasoning performance, making process supervision a useful signal for guiding inference~\cite{malik2025rewardbench}. However, directly using PRMs at test time is expensive: candidate reasoning steps must be \textit{first generated} and then evaluated by a separate verifier. This creates substantial overhead, especially for long reasoning traces or multi-sample decoding. {\ourmethod} addresses this bottleneck by distilling PRM supervision into a lightweight verifier that predicts step quality directly from hidden states, enabling verifier-guided control without explicit text-level verification at inference time.

\section{Proposed Method: {\ourmethod}}
\label{sec:method}
Given a target LLM $\mathcal{M}$ and a small construction set
$\mathcal{D}_{\text{train}}$, {\ourmethod} learns an external latent verifier
$\mathcal{V}_{\phi}$ that guides test-time model steering. The framework
has two phases. Offline, we generate reasoning traces, segment them into
thought units, extract hidden states at thought boundaries, and construct
steering vectors for different reasoning modes. We also distill PRM-provided
step-level scores into a lightweight verifier over the extracted hidden states. Online, the
verifier scores candidate latent interventions and selects the steering action to continue generation. Figure~\ref{fig:atlas_overview} illustrates the overall pipeline.

\subsection{Problem Formulation}
\label{sec:thought_states_actions}
Given an input problem \(x\), the model generates a reasoning trace
\(\mathcal{R}=(u_1,\ldots,u_T)\), where each \(u_t\) is a contiguous reasoning unit, or \emph{thought}. In our implementation, thoughts are separated by paragraph boundaries, although the framework is compatible with other boundary detectors. Let \(\tau_t\) denote the token position ending thought \(u_t\). At a designated transformer layer \(\ell\), we extract the boundary hidden state:
\begin{equation}
    z_t^{(\ell)} = h_{\tau_t}^{(\ell)} \in \mathbb{R}^d .
\end{equation}
This state summarizes the reasoning trajectory \textit{up to} step \(t\) and serves as the latent state for action selection.

{\ourmethod} uses a discrete action space
\(\mathcal{A}=\{\varnothing,E,R,T\}\), where \(\varnothing\) denotes no
intervention, and \(E\), \(R\), and \(T\) denote execution-, reflection-, and
transition-oriented steering actions, respectively. Each action \(a\in
\mathcal{A}\) is associated with an action vector \(v_a^{(\ell)}\) constructed
offline, with \(v_{\varnothing}^{(\ell)}=\mathbf{0}\). Selecting action \(a_t\) produces the intervened state:
\begin{equation}
    \tilde{z}_t^{(\ell)}
    =
    z_t^{(\ell)}+\alpha v_{a_t}^{(\ell)},
    \label{eq:intervention}
\end{equation}
where \(\alpha\) controls the intervention strength. The goal is to select, at
each thought boundary, the action that best improves subsequent reasoning, while
avoiding unnecessary generation.

\subsection{Offline Steering Vector Construction}
\label{offline_steering_vector_construction}
\paragraph{Thought Segmentation and Hidden-State Extraction.} For each problem in $\mathcal{D}_{\text{train}}$, we run the target model
$\mathcal{M}$ to generate a full reasoning trace. We segment the trace into
thoughts using paragraph boundaries, implemented as the delimiter
``\texttt{\textbackslash n\textbackslash n}''. Let $\tau_t$ denote the token
position ending thought $u_t$. We extract the residual-stream activation after
transformer block $\ell$ at this boundary:
\begin{equation}
    z_t^{(\ell)} = h_{\tau_t}^{(\ell)} .
\end{equation}
This boundary representation summarizes the reasoning trajectory up to step
$t$ and serves as the latent state used for both steering-vector construction
and verifier training. To obtain weak thought-type labels, we categorize each thought as \emph{execution}, \emph{reflection}, or \emph{transition} using lightweight keyword-based heuristics following prior work. These labels are used only to construct steering directions, not as ground-truth semantic annotations. The
labeling rules are described in Appendix~\ref{thoughts_classification}, and
their robustness is evaluated in Appendix~\ref{sec:label_robustness}.

\paragraph{Steering Vector Construction.}
We construct steering directions by contrasting hidden-state representations of different thought categories. Let \(\mathcal{I}_c\) denote the set of example step pairs \((x,t)\) whose thought \(u_t\) is labeled with reasoning mode \(c\in\{E,R,T\}\), and let \(\mathcal{I}\) denote all thought boundaries. For each mode, we first compute the average hidden representation:
\begin{equation}
    \bar{z}_c^{(\ell)}
    =
    \frac{1}{|\mathcal{I}_c|}
    \sum_{(x,t)\in \mathcal{I}_c}
    z_t^{(\ell)}(x).
\end{equation}
We also compute the complementary-mode average
\begin{equation}
    \bar{z}_{\neg c}^{(\ell)}
    =
    \frac{1}{|\mathcal{I}\setminus \mathcal{I}_c|}
    \sum_{(x,t)\in \mathcal{I}\setminus \mathcal{I}_c}
    z_t^{(\ell)}(x).
\end{equation}
The action vector for mode \(c\) is then defined as the contrastive direction
between the target mode and its complement:
\begin{equation}
    v_c^{(\ell)}
    =
    \bar{z}_c^{(\ell)}
    -
    \bar{z}_{\neg c}^{(\ell)}.
    \label{eq:action_vector_construction}
\end{equation}
Together with \(v_{\varnothing}^{(\ell)}=\mathbf{0}\), this yields the
candidate action vectors
\(\{v_{\varnothing}^{(\ell)},v_E^{(\ell)},v_R^{(\ell)},v_T^{(\ell)}\}\).
The execution action \(v_E^{(\ell)}\) corresponds to the SEAL-style
execution-versus-non-execution contrast, while \(v_R^{(\ell)}\) and
\(v_T^{(\ell)}\) extend the same contrastive mechanism to reflection and
transition. These action vectors are computed once offline and reused during
inference.

\subsection{Lightweight Latent Verifier Training}
The latent verifier is a \textit{compact MLP with only two hidden-layer} that maps \(z_t^{(\ell)} \in \mathbb{R}^{d_{\mathcal{M}}}\) to a quality score in \([0,1]\), where \(d_{\mathcal{M}}\) is the hidden dimension of the target model. Across all settings, we use the architecture
\(d_{\mathcal{M}}\!\rightarrow\!256\!\rightarrow\!128\!\rightarrow\!1\), with ReLU activations, dropout \(0.2\), and a sigmoid output. For each target model, we train the verifier on hidden states extracted at the corresponding
intervention layer, using PRM-provided continuous step-quality scores as supervision. We split the verifier data into 6/2/2 train/validation/test partitions and optimized mean squared error on the PRM scores. This design keeps the verifier small across target models: the verifier has 426K parameters for R1-Distill-Qwen-1.5B and 1.34M parameters for 32B-scale models, with detailed training costs reported in Appendix~\ref{appendix:verifier_training_cost}.

\subsection{Adaptive Steering Policy}
Because downstream correctness is unavailable during generation, ATLAS uses the latent verifier \(\mathcal{V}_{\phi}\) to estimate the quality of candidate interventions before generation continues. Specifically, at each thought boundary, {\ourmethod} evaluates all candidate actions in
$\mathcal{A}$. For each action \(a\), it forms a candidate intervened state
\(z_t^{(\ell)} + \alpha v_a^{(\ell)}\) and scores it with the verifier.
The selected action is
\begin{equation}
    a_t^{\star} 
    = \arg\max_{a \in \mathcal{A}}
    \mathcal{V}_{\phi}
    \left(
    z_t^{(\ell)} + \alpha v_a^{(\ell)}
    \right).
    \label{eq:adaptive_policy}
\end{equation}
The chosen action is then applied according to Eq.~\ref{eq:intervention}, and
generation continues until the next thought boundary. This procedure enables the model to continue execution when the reasoning trajectory is reliable, trigger reflection when uncertainty or error is detected, and transition when the current trajectory appears unproductive. Unless otherwise stated, we use $\alpha=1.0$ and intervene at middle transformer layers, which we find most effective empirically. Layer sensitivity is analyzed in Appendix~\ref{layer_ablation_study}.

\paragraph{Algorithm Detail.}
Algorithm~\ref{alg:atlas_inference} presents the inference procedure of {\ourmethod}. At each thought boundary, the verifier scores candidate interventions induced by the available steering actions, and {\ourmethod} applies the action with the highest predicted quality before continuing generation.
\begin{algorithm}[t]
\footnotesize
\caption{{\ourmethod} Adaptive Inference}
\label{alg:atlas_inference}
\begin{algorithmic}[1]
\Require $\mathcal{M}$, input $x$, verifier $\mathcal{V}_{\phi}$, action vectors $\{v_{\varnothing}^{(\ell)},v_E^{(\ell)},v_R^{(\ell)},v_T^{(\ell)}\}$, layer $\ell$, strength $\alpha$
\Ensure Generated solution $y$
\State $\mathcal{A}\leftarrow\{\varnothing,E,R,T\}$, $v_{\varnothing}^{(\ell)}\leftarrow\mathbf{0}$, $y\leftarrow\emptyset$
\While{not terminated}
    \State Generate until the next thought boundary; append tokens to $y$
    \If{generation terminates}
        \State \textbf{break}
    \EndIf
    \State Extract boundary state $z_t^{(\ell)}$
    \For{$a \in \mathcal{A}$}
        \State $r_a \leftarrow \mathcal{V}_{\phi}\!\left(z_t^{(\ell)}+\alpha v_a^{(\ell)}\right)$
    \EndFor
    \State $a_t^{\star} \leftarrow \arg\max_{a\in\mathcal{A}} r_a$
    \State Apply $z_t^{(\ell)} \leftarrow z_t^{(\ell)}+\alpha v_{a_t^{\star}}^{(\ell)}$
\EndWhile
\State \Return $y$
\end{algorithmic}
\end{algorithm}

\paragraph{Computational Cost.}
The steering vectors are computed offline. During inference, {\ourmethod} adds only four forward passes through a small MLP per thought boundary, one for each candidate action. For verifier hidden width \(m\) and model hidden dimension \(d\), this cost is bounded by \(O(|\mathcal{A}|dm)\) per boundary, plus a vector addition for the selected intervention. No additional LLM decoding passes or PRM calls are required. The verifier is also lightweight to train: even for 32B-scale target models, it contains only 1.34M parameters and requires approximately one minute for 100 epochs on a single H100 GPU; detailed architecture and training-cost statistics are provided in Appendix~\ref{appendix:verifier_training_cost}. This makes latent verification substantially cheaper than text-level adaptive verification, which must generate candidate reasoning steps and score them with an external reward model.

\section{Experimental Setup}
\label{sec:experimental_setup}
\paragraph{Datasets, Models, and Metrics.}
We evaluate {\ourmethod} on five mathematical reasoning benchmarks, GSM8K~\citep{cobbe2021training}, MATH~\citep{hendrycks2021measuring}, AMC2023~\citep{amc2023}, AIME2024, and AIME2025~\citep{sun2025challenging}, and one coding benchmark, LiveCodeBench~\citep{jain2025livecodebench}. 
These benchmarks cover grade-school arithmetic, competition-level mathematics, olympiad-style reasoning, and temporally held-out code generation. 

We experiment with DeepSeek-R1-Distill-Qwen-1.5B/7B/32B, denoted as R1-Distill-1.5B/7B/32B~\citep{guo2025deepseek}, and QwQ-32B-Preview~\citep{qwq-32b-preview,qwen2}.  All methods use the same instruction prompt and a maximum generation length of 8,192 tokens. We report final-answer accuracy (\textit{Acc.}) and average generated tokens (\textit{\#Tok}), excluding prompt tokens. Higher accuracy and lower token usage indicate a better accuracy--efficiency trade-off. Additional benchmark and evaluation details are provided in Appendix~\ref{appendix:experimental_details}.

\paragraph{Verifier Supervision.}
To train the latent verifier, we generate reasoning traces from a construction set $\mathcal{D}_{\text{train}}$ using the target model and segment each trace into thought units. For every thought boundary, we extract the hidden state $z_t^{(\ell)}$ at the intervention layer. We obtain step-level supervision using Math-Shepherd\footnote{\url{https://huggingface.co/peiyi9979/math-shepherd-mistral-7b-prm}}, a 7B process reward model that assigns quality scores to intermediate steps in mathematical solutions~\cite{wang2024math}. This yields training pairs $(z_t^{(\ell)}, q_t)$, where $q_t \in [0,1]$ is the PRM-provided quality score. We split these pairs into training, validation, and test partitions with a 6:2:2 ratio. The PRM is used only for offline verifier supervision and for the text-level adaptive baseline; {\ourmethod} with latent verification does not call the PRM during inference.

\paragraph{Baselines.} We compare against three groups of methods.
\textit{(i) Vanilla} uses the base model without any steering intervention.
\textit{(ii) Fixed steering} applies a single non-adaptive intervention throughout generation. This group includes ASC~\cite{azizi2025activation}, which uses activation steering to encourage concise reasoning; SEAL~\cite{chen2025seal}, which modifies latent representations associated with reasoning modes; and CREST~\cite{zhang2025understanding}, which applies calibrated latent-space rotations to reduce inefficient reasoning patterns such as overthinking and backtracking. \textit{(iii) Adaptive steering} includes {\ourmethod}, our main method, which selects steering actions using the lightweight latent verifier, and {\ourmethod}-T, a text-level verifier variant that selects actions using PRM-based verification of generated reasoning steps. Throughout the paper, {\ourmethod} refers to the latent-verifier version of our method unless otherwise specified.

\section{Main Results}
\label{main_results}
We evaluate {\ourmethod} under two settings. In the \emph{in-domain} setting, steering vectors and the latent verifier are constructed from development data drawn from the same benchmark family as the evaluation benchmark. In the \emph{cross-domain} setting, the same controller is constructed from a source benchmark and transferred to target benchmarks without using target-domain examples. This distinction lets us separately evaluate benchmark-matched adaptation and transfer under distribution shift. Unless otherwise specified, {\ourmethod} refers to the latent-verifier version of our method, while {\ourmethod}-T denotes the text-level verifier variant.

\begin{table}[t]
\centering
\scriptsize
\setlength{\tabcolsep}{1.8pt}
\renewcommand{\arraystretch}{1.08}
\resizebox{\columnwidth}{!}{%
\begin{tabular}{l c c c c c}
\toprule
\multicolumn{2}{c}{\textbf{Model}}
& \multicolumn{2}{c}{\textbf{MATH}}
& \multicolumn{2}{c}{\textbf{GSM8K}} \\
\cmidrule(lr){3-4} \cmidrule(lr){5-6}
& & Acc. $\uparrow$ & Tok. $\downarrow$
  & Acc. $\uparrow$ & Tok. $\downarrow$ \\
\midrule

\textit{Vanilla} & \multirow{5}{*}{\rotatebox{90}{\textit{R1-Distill-1.5B}}}
& \basecell{73.76} & \basecell{3941}
& \basecell{79.30} & \basecell{2390} \\
\textit{SEAL} &
& \accgain{\underline{79.78}}{8.4} & \tokgain{\underline{3034}}{23.0}
& \accgain{\underline{82.41}}{3.9} & \tokgain{\underline{1438}}{39.8} \\
\textit{ASC} &
& \accgain{73.98}{0.3} & \tokgain{3252}{17.5}
& \accgain{79.61}{0.4} & \tokgain{2267}{5.1} \\
\textit{CREST} &
& \accloss{69.06}{6.2} & \tokloss{4016}{1.9}
& \accloss{75.44}{4.9} & \tokgain{2142}{10.4} \\
\textit{ATLAS} &
& \oursaccgain{82.28}{11.6} & \ourstokgain{2754}{30.1}
& \oursaccgain{85.37}{7.7} & \ourstokgain{1316}{44.9} \\
\midrule

\textit{Vanilla} & \multirow{5}{*}{\rotatebox{90}{\textit{R1-Distill-7B}}}
& \basecell{86.34} & \basecell{3395}
& \basecell{89.23} & \basecell{841} \\
\textit{SEAL} &
& \accgain{\underline{88.96}}{3.0} & \tokgain{3050}{10.2}
& \accloss{88.55}{0.8} & \tokgain{\underline{655}}{22.1} \\
\textit{ASC} &
& \neutralcell{86.38}{0.0} & \tokgain{\underline{3026}}{10.9}
& \accloss{86.50}{3.1} & \tokgain{795}{5.5} \\
\textit{CREST} &
& \accgain{86.48}{0.2} & \tokgain{3130}{7.8}
& \accgain{\underline{89.38}}{0.2} & \tokloss{1243}{47.8} \\
\textit{ATLAS} &
& \oursaccgain{{90.68}}{5.0} & \ourstokgain{2895}{14.7}
& \oursaccgain{\textbf{89.61}}{0.4} & \ourstokgain{637}{24.3} \\
\midrule

\textit{Vanilla} & \multirow{5}{*}{\rotatebox{90}{\textit{R1-Distill-32B}}}
& \basecell{91.54} & \basecell{2413}
& \basecell{92.87} & \basecell{444} \\
\textit{SEAL} &
& \accgain{\underline{92.42}}{1.0} & \tokgain{2030}{15.9}
& \accgain{92.95}{0.1} & \tokgain{\underline{436}}{1.8} \\
\textit{ASC} &
& \accgain{91.60}{0.1} & \tokgain{\underline{2014}}{16.5}
& \accgain{\underline{93.03}}{0.2} & \tokloss{493}{11.0} \\
\textit{CREST} &
& \accgain{91.60}{0.1} & \tokgain{2156}{10.6}
& \accgain{92.95}{0.1} & \tokloss{1143}{157.4} \\
\textit{ATLAS} &
& \oursaccgain{93.04}{1.6} & \ourstokgain{1896}{21.4}
& \oursaccgain{\textbf{93.56}}{0.7} & \ourstokgain{427}{3.8} \\
\midrule

\textit{Vanilla} & \multirow{5}{*}{\rotatebox{90}{\textit{QwQ-32B-Pre}}}
& \basecell{90.02} & \basecell{2114}
& \basecell{\underline{94.54}} & \basecell{687} \\
\textit{SEAL} &
& \accgain{\underline{90.20}}{0.2} & \tokgain{2043}{3.4}
& \accloss{94.39}{0.2} & \tokgain{\underline{508}}{26.1} \\
\textit{ASC} &
& \accgain{90.10}{0.1} & \tokgain{\underline{2015}}{4.7}
& \accloss{94.24}{0.3} & \tokgain{539}{21.5} \\
\textit{CREST} &
& \accgain{{90.16}}{0.2} & \tokgain{2032}{3.9}
& \accloss{94.31}{0.2} & \tokloss{1034}{50.5} \\
\textit{ATLAS} &
& \oursaccgain{90.42}{0.4} & \ourstokgain{1959}{7.3}
& \oursaccgain{{94.84}}{0.3} & \ourstokgain{498}{27.5} \\
\bottomrule
\end{tabular}%
}
\vspace{-3pt}
\caption{In-domain reasoning performance on MATH and GSM8K. We report final-answer accuracy (Acc., \%) and average generated tokens (Tok.). Values in parentheses indicate relative change from the Vanilla baseline within the same model and benchmark. For Acc., positive values indicate improvement; for Tok., negative values indicate token reduction. Blue cells denote improvements over Vanilla, red cells denote degradation, gray cells denote Vanilla or no change, and blue-shaded rows highlight ATLAS. The best and second-best results are marked in \textbf{bold} and \underline{underline}, respectively.}
\label{tab:indomain_performance}
\vspace{-6pt}
\end{table}

\begin{table*}[t]
\centering
\scriptsize
\setlength{\tabcolsep}{1.2pt}
\renewcommand{\arraystretch}{1.08}
\resizebox{\textwidth}{!}{%
\begin{tabular}{l c c c c c c c c c c c}
\toprule
\multicolumn{2}{c}{\textbf{Model}}
& \multicolumn{2}{c}{\textbf{GSM8K}}
& \multicolumn{2}{c}{\textbf{AIME2024}}
& \multicolumn{2}{c}{\textbf{AMC2023}}
& \multicolumn{2}{c}{\textbf{AIME2025}}
& \multicolumn{2}{c}{\textbf{LiveCodeBench}} \\
\cmidrule(lr){3-4} \cmidrule(lr){5-6} \cmidrule(lr){7-8}
\cmidrule(lr){9-10} \cmidrule(lr){11-12}
& & Acc. $\uparrow$ & Tok. $\downarrow$
  & Acc. $\uparrow$ & Tok. $\downarrow$
  & Acc. $\uparrow$ & Tok. $\downarrow$
  & Acc. $\uparrow$ & Tok. $\downarrow$
  & Acc. $\uparrow$ & Tok. $\downarrow$ \\
\midrule

\textit{Vanilla} & \multirow{5}{*}{\rotatebox{90}{\textit{R1-Distill-1.5B}}}
& \basecell{79.30} & \basecell{2390}
& \basecell{20.00} & \basecell{7396}
& \basecell{47.50} & \basecell{5241}
& \basecell{10.00} & \basecell{7893}
& \basecell{19.00}	 & \basecell{8268.32} \\
\textit{SEAL} &
& \accgain{\underline{83.32}}{5.1} & \tokgain{\underline{1432}}{40.1}
& \accgain{\underline{23.33}}{16.7} & \tokgain{\underline{6820}}{7.8}
& \accgain{52.50}{10.5} & \tokgain{\underline{5111}}{2.5}
& \accgain{\underline{20.00}}{100.0} & \tokgain{\underline{6763}}{14.3}
& \accgain{\underline{29.25}}{53.9} & \tokgain{\underline{6978}}{15.6}\\
\textit{ASC} &
& \accloss{77.41}{2.4} & \tokloss{2647}{10.8}
& \accloss{10.00}{50.0} & \tokloss{7744}{4.7}
& \accloss{42.50}{10.5} & \tokloss{5590}{6.7}
& \accgain{\textbf{23.33}}{133.3} & \tokgain{7082}{10.3}
& \accloss{18.00}{5.3} & \tokloss{8520}{3.0} \\
\textit{CREST} &
& \accloss{75.28}{5.1} & \tokgain{1799}{24.7}
& \accgain{\underline{23.33}}{16.7} & \tokgain{7063}{4.5}
& \accgain{\underline{57.50}}{21.1} & \tokloss{5349}{2.1}
& \accgain{\underline{20.00}}{100.0} & \tokgain{7021}{11.0}
& \accgain{27.50}{44.7} & \tokloss{9050}{9.5} \\
\textit{ATLAS} &
& \oursaccgain{84.53}{6.6} & \ourstokgain{1171}{51.0}
& \oursaccgain{33.33}{66.7} & \ourstokgain{6304}{14.8}
& \oursaccgain{65.00}{36.8} & \ourstokgain{3837}{26.8}
& \oursaccgain{{23.33}}{133.3} & \ourstokgain{{6337}}{19.7}
& \oursaccgain{\textbf{32.00}}{68.4} & \ourstokgain{\textbf{6216}}{24.8}\\
\midrule

\textit{Vanilla} & \multirow{5}{*}{\rotatebox{90}{\textit{R1-Distill-7B}}}
& \basecell{89.23} & \basecell{841}
& \basecell{40.00} & \basecell{7269}
& \basecell{82.50} & \basecell{4458}
& \basecell{\underline{33.33}} & \basecell{6891}
& \basecell{28.00} & \basecell{7050}  \\
\textit{SEAL} &
& \accloss{88.63}{0.7} & \tokgain{\underline{657}}{21.9}
& \accgain{\underline{46.67}}{16.7} & \tokgain{\underline{6222}}{14.4}
& \neutralcell{\underline{82.50}}{0.0} & \tokgain{\underline{3822}}{14.3}
& \accloss{20.00}{40.0} & \tokgain{6773}{1.7}
& \accgain{\underline{35.00}}{25.0} & \tokgain{\underline{6220}}{11.8} \\
\textit{ASC} &
& \accloss{88.70}{0.6} & \tokloss{848}{0.8}
& \accloss{36.67}{8.3} & \tokgain{7001}{3.7}
& \accloss{70.00}{15.2} & \tokloss{4624}{3.7}
& \accloss{26.67}{20.0} & \tokloss{7124}{3.4}
& \accloss{26.00}{7.1} & \tokloss{7310}{3.7} \\
\textit{CREST} &
& \accgain{\underline{89.84}}{0.7} & \tokloss{1604}{90.7}
& \neutralcell{40.00}{0.0} & \tokgain{6835}{6.0}
& \accloss{72.50}{12.1} & \tokloss{4743}{6.4}
& \accloss{23.33}{30.0} & \tokgain{\underline{6882}}{0.1}
& \accgain{34.50}{23.2} & \tokloss{8120}{15.2} \\
\textit{ATLAS} &
& \oursaccgain{{89.61}}{0.4} & \ourstokgain{611}{27.4}
& \oursaccgain{{56.67}}{41.7} & \ourstokgain{5694}{21.7}
& \oursaccgain{87.50}{6.1} & \ourstokgain{3749}{15.9}
& \oursaccgain{\textbf{40.00}}{20.0} & \ourstokgain{6010}{12.8}
& \oursaccgain{{38.50}}{37.5} & \ourstokgain{{5890}}{16.5} \\
\midrule

\textit{Vanilla} & \multirow{5}{*}{\rotatebox{90}{\textit{R1-Distill-32B}}}
& \basecell{92.87} & \basecell{\underline{444}}
& \basecell{40.00} & \basecell{6654}
& \basecell{80.00} & \basecell{4221}
& \basecell{\underline{36.67}} & \basecell{6705}
& \basecell{40.00} & \basecell{5450} \\
\textit{SEAL} &
& \accloss{92.04}{0.9} & \tokgain{\textbf{425}}{4.3}
& \accgain{\underline{46.67}}{16.7} & \tokgain{6110}{8.2}
& \accgain{\textbf{90.00}}{12.5} & \tokgain{3442}{18.5}
& \neutralcell{\underline{36.67}}{0.0} & \tokgain{\underline{6261}}{6.6}
& \accgain{\underline{45.00}}{12.5} & \tokgain{\underline{4900}}{10.1} \\
\textit{ASC} &
& \accloss{91.74}{1.2} & \tokloss{454}{2.3}
& \accloss{36.67}{8.3} & \tokgain{6432}{3.3}
& \accloss{75.00}{6.3} & \tokgain{4217}{0.1}
& \accloss{30.00}{18.2} & \tokloss{6892}{2.8}
& \accloss{37.00}{7.5} & \tokloss{5600}{2.8} \\
\textit{CREST} &
& \accgain{\underline{93.03}}{0.2} & \tokloss{1132}{155.0}
& \accgain{\underline{46.67}}{16.7} & \tokgain{\underline{6082}}{8.6}
& \accgain{\underline{82.50}}{3.1} & \tokgain{\underline{3427}}{18.8}
& \neutralcell{\underline{36.67}}{0.0} & \tokgain{6632}{1.1}
& \accgain{44.50}{11.2} & \tokloss{6200}{13.8}\\
\textit{ATLAS} &
& \oursaccgain{{93.63}}{0.8} & \ourstokgain{{425}}{4.3}
& \oursaccgain{{60.00}}{50.0} & \ourstokgain{{5442}}{18.2}
& \accgain{\underline{82.50}}{3.1} & \ourstokgain{{3134}}{25.8}
& \oursaccgain{\textbf{43.33}}{18.2} & \ourstokgain{{5987}}{10.7}
& \oursaccgain{\textbf{49.50}}{23.8} & \ourstokgain{\textbf{4380}}{19.6}\\
\midrule

\textit{Vanilla} & \multirow{5}{*}{\rotatebox{90}{\textit{QwQ-32B-Pre}}}
& \basecell{94.54} & \basecell{687}
& \basecell{33.33} & \basecell{5977}
& \basecell{\underline{82.50}} & \basecell{3778}
& \basecell{33.33} & \basecell{6048}
& \basecell{43.00} & \basecell{5000}  \\
\textit{SEAL} &
& \accloss{92.80}{1.8} & \tokloss{\underline{696}}{1.3}
& \accgain{\underline{43.33}}{30.0} & \tokgain{\textbf{5465}}{8.6}
& \accloss{80.00}{3.0} & \tokgain{\textbf{3352}}{11.3}
& \accloss{30.00}{10.0} & \tokgain{\underline{5756}}{4.8}
& \accgain{\underline{47.00}}{9.3} & \tokgain{\underline{4550}}{9.0} \\
\textit{ASC} &
& \accloss{94.01}{0.6} & \tokloss{812}{18.2}
& \accloss{30.00}{10.0} & \tokgain{5902}{1.3}
& \accloss{77.50}{6.1} & \tokloss{3792}{0.4}
& \accloss{23.33}{30.0} & \tokgain{5932}{1.9}
& \accloss{41.00}{4.7} & \tokloss{5200}{4.0}\\
\textit{CREST} &
& \accgain{\underline{95.00}}{0.5} & \tokloss{1242}{80.8}
& \accgain{40.00}{20.0} & \tokgain{5694}{4.7}
& \neutralcell{\underline{82.50}}{0.0} & \tokgain{\underline{3521}}{6.8}
& \accgain{\underline{36.67}}{10.0} & \tokloss{6433}{6.4}
& \accgain{46.50}{8.1} & \tokloss{5650}{13.0}\\
\textit{ATLAS} &
& \oursaccgain{\underline{95.00}}{0.5} & \ourstokgain{{449}}{34.6}
& \oursaccgain{{56.67}}{70.0} & \tokgain{\underline{5621}}{6.0}
& \oursaccgain{{85.00}}{3.0} & \tokgain{\underline{3521}}{6.8}
& \oursaccgain{{40.00}}{20.0} & \ourstokgain{\textbf{5732}}{5.2}
& \oursaccgain{\textbf{51.50}}{19.8} & \ourstokgain{\textbf{4150}}{17.0}\\
\bottomrule
\end{tabular}%
}
\vspace{-3pt}
\caption{Cross-domain performance. We report final-answer accuracy (Acc., \%) and average generated tokens (Tok.). Values in parentheses indicate relative change from Vanilla baseline within the same model, benchmark. For Acc., positive values indicate improvement. For Tok., negative values mean token reduction. Blue cells: improvements over Vanilla; Red cells: degradation; Gray cells: Vanilla or no change.
}
\label{tab:crossdomain_performance}
\vspace{-5pt}
\end{table*}

\subsection{In-Domain Performance}
We show overall performance for the \textbf{in-domain} setting in Table~\ref{tab:indomain_performance}. ATLAS consistently improves accuracy while reducing generated tokens across all model--benchmark pairs. The gains are largest for smaller models: on R1-Distill-1.5B, ATLAS improves MATH accuracy from 73.76\% to 82.28\% while reducing token usage by 30.1\%, and improves GSM8K accuracy from 79.30\% to 85.37\% while reducing token usage by 44.9\%. Fixed steering baselines are less stable. SEAL often reduces token usage but can underperform on accuracy, while ASC and CREST show mixed behavior across models and datasets. In contrast, ATLAS adapts the intervention at each reasoning boundary, allowing it to preserve useful reasoning steps while suppressing redundant or unproductive generation. These results show that the benefit comes not merely from applying a steering vector, but from selecting the intervention based on the current latent reasoning state.

\subsection{Out-of-Domain Transferability}
Table~\ref{tab:crossdomain_performance} evaluates ATLAS on more challenging
mathematical benchmarks and LiveCodeBench. ATLAS maintains favorable
accuracy--efficiency trade-offs beyond the in-domain setting. On R1-Distill-1.5B,
ATLAS improves AIME2024 accuracy from 20.00\% to 33.33\% while reducing token
usage by 14.8\%, and improves AMC2023 accuracy from 47.50\% to 65.00\% with a
26.8\% token reduction. The gains are especially pronounced on contest-style
benchmarks, where fixed execution-oriented steering is often insufficient and
the model benefits from adaptively switching among execution, reflection, and
transition actions. The results also reveal a scale-dependent pattern. Smaller models benefit substantially from latent adaptive steering, likely because they require more frequent correction and strategy shifts. For larger models, accuracy gains are smaller on easier benchmarks where the base model is already strong, but {\ourmethod} still often reduces token usage and improves performance on harder tasks such as AIME and LiveCodeBench. Overall, these results indicate that latent verification provides a transferable control signal for adaptive reasoning.

\subsection{Latent Verification Efficiently Approximates Text-Level Verification}
Table~\ref{tab:atlas_t_vs_l_summary} compares {\ourmethod} with {\ourmethod}-T. {\ourmethod}-T uses text-level PRM verification to select steering actions, while {\ourmethod} replaces this expensive procedure with a lightweight verifier over hidden states. Across in-domain benchmarks, {\ourmethod} matches or slightly outperforms {\ourmethod}-T on average while using fewer generated tokens. Under transfer, {\ourmethod} remains competitive with {\ourmethod}-T, with a small accuracy gap on larger models but comparable token efficiency. These results support the central design choice of {\ourmethod}: hidden-state quality signals are sufficient to recover most of the benefit of text-level verification, while avoiding additional candidate generation and inference-time PRM calls. Full per-benchmark {\ourmethod}-T results are provided in
Appendix~\ref{appendix:atlas_t_results}.

\begin{table}[t]
\centering
\footnotesize
\setlength{\tabcolsep}{1.5pt}
\resizebox{\columnwidth}{!}{%
\begin{tabular}{clrrrrrr}
\toprule
\multirow{2}{*}{\textbf{Setting}}
& \multirow{2}{*}{\textbf{Model}}
& \multicolumn{2}{c}{\textbf{Base}}
& \multicolumn{2}{c}{\textbf{ATLAS(T)}} 
& \multicolumn{2}{c}{\textbf{ATLAS}} \\
\cmidrule(lr){3-4} \cmidrule(lr){5-6} \cmidrule(lr){7-8}
& & Acc. $\uparrow$ & Tok. $\downarrow$
  & Acc. $\uparrow$ & Tok. $\downarrow$
  & Acc. $\uparrow$ & Tok. $\downarrow$ \\
\midrule

\multirow{4}{*}{\rotatebox[origin=c]{90}{\scriptsize\textbf{In-dom.}}}
& R1-Distill-1.5B & 76.53 & 3166 & 83.75 & 2118 & 83.83 & 2035 \\
& R1-Distill-7B  & 87.79 & 2118 & 90.08 & 1824 & 90.15 & 1766 \\
& R1-Distill-32B  & 92.21 & 1429 & 93.25 & 1205 & 93.30 & 1162 \\
& QwQ-32B-Pre   & 92.28 & 1401 & 92.74 & 1237 & 92.63 & 1228 \\
\midrule

\multirow{4}{*}{\rotatebox[origin=c]{90}{\scriptsize\textbf{Cross-dom.}}}
& R1-Distill-1.5B  & 35.16 & 6238 & 46.93 & 4712 & 47.64 & 4773 \\
& R1-Distill-7B  & 54.61 & 5302 & 62.09 & 4438 & 62.46 & 4391 \\
& R1-Distill-32B  & 57.91 & 4695 & 68.09 & 3841 & 65.79 & 3874 \\
& QwQ-32B-Pre  & 57.34 & 4298 & 67.41 & 3897 & 65.63 & 3895 \\
\bottomrule
\end{tabular}%
}
\caption{Comparison between base decoding, text-level verification, and latent verification. We report average accuracy and generated tokens across in-domain benchmarks and cross-domain benchmarks. Base denotes vanilla decoding without steering.}
\label{tab:atlas_t_vs_l_summary}
\vspace{-10pt}
\end{table}

\section{Analysis and Discussion}
\label{sec:analysis}
\begin{figure}[tb]
    \centering
    \includegraphics[width=0.49\textwidth]{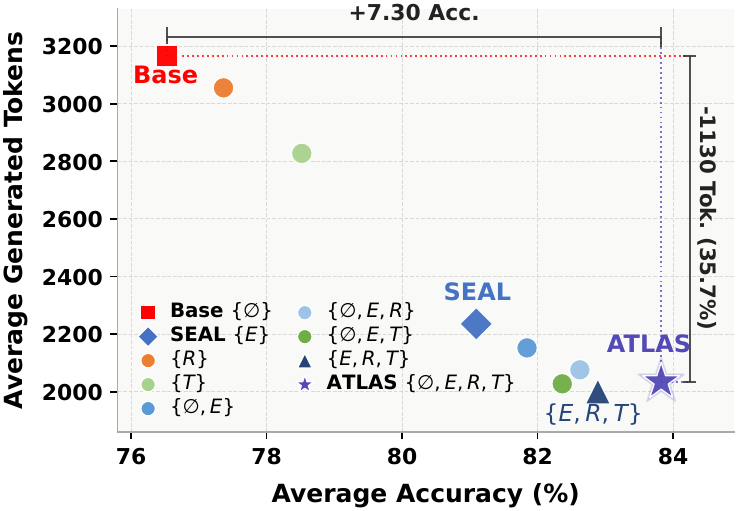}
    \caption{Accuracy--efficiency trade-off of different action sets on R1-Distill-Qwen-1.5B, averaged across MATH and GSM8K. Each point corresponds to an action-set configuration, with higher accuracy and lower generated tokens indicating better reasoning efficiency.}
    \label{fig:action_set_tradeoff}
    \vspace{-5pt}
\end{figure}

\noindent \textbf{Do Adaptive Steering Actions Matter?}
Figure~\ref{fig:action_set_tradeoff} summarizes the average accuracy--efficiency trade-off across the two benchmarks. 
The execution-only setting $\{E\}$, which corresponds to a static SEAL-style policy, improves over the no-intervention baseline, confirming that execution-oriented steering is a useful component.  However, it is much less effective than adaptive strategy.  Using only reflection or transition performs worse than execution, suggesting that corrective or exploratory actions are insufficient without execution-oriented progress.  Adding no-intervention action $\varnothing$ further improves the trade-off, indicating that abstaining from unnecessary perturbations helps avoid over-steering.  Finally, incorporating reflection and transition actions provides additional gains, with the full action set $\{\varnothing,E,R,T\}$ achieving the highest average accuracy while reducing generated tokens by 35.7\% compared with the base model.  These results show that the gains of {\ourmethod} come not merely from stronger execution steering, but from adaptively selecting among complementary reasoning modes. Detailed, full per-benchmark results are provided in Appendix~\ref{appendix_action_set_study}

\vspace{5pt}
\noindent \textbf{Policy Dynamics versus Model Scale.}
Table~\ref{tab:steering_distribution} reports the distribution of selected actions across model sizes. Smaller models select reflection and transition more frequently, while larger models rely more heavily on execution. This suggests that weaker models benefit more from corrective and exploratory interventions, whereas stronger models more often maintain a productive execution trajectory. The non-trivial use of all four actions further indicates that {\ourmethod} does not collapse to a fixed steering policy.

\vspace{5pt}
\noindent \textbf{Stability Under Perturbation.}
We evaluate whether the latent verifier remains reliable under small hidden-state shifts, since adaptive steering scores perturbed candidate states during inference. Because steering vectors are derived from mean differences between reasoning modes, the interventions are expected to remain within a locally meaningful representation subspace rather than shifting off-manifold. Empirically, we add Gaussian perturbations to held-out hidden states and measure the absolute prediction change, $|\Delta \mathrm{Score}|$, using variance estimated from the hidden-state distribution, $\sigma^2{=}3.07$. As shown in Table~\ref{tab:verifier_stability}, prediction changes remain insignificant ($<$0.01) for noise scales up to 0.25 and stay small even under larger perturbations, suggesting that the verifier provides a robust ranking signal for adaptive steering. A qualitative example is provided in Appendix~\ref{appendix:qualitative_correlation}.

\begin{table}[tb!]
\footnotesize
\setlength{\tabcolsep}{1pt}
\centering
\begin{tabular}{lcccc}
\toprule
\textbf{Model} & \textbf{Execution} & \textbf{Reflection} & \textbf{Transition} & \textbf{Neutral} \\
\midrule
R1-Distill-1.5B  & 37.6\% & 27.1\% & 21.2\% & 14.1\% \\
R1-Distill-7B    & 49.4\% & 15.3\% & 18.8\% & 16.5\% \\
R1-Distill-32B   & 62.3\% & 10.4\% & 14.3\% & 13.0\% \\
QwQ-32B-Pre      & 58.8\% & 12.9\% & 15.3\% & 13.0\% \\
\bottomrule
\end{tabular}
\caption{Steering mode distribution across model sizes.}
\label{tab:steering_distribution}
\vspace{-10pt}
\end{table}

\vspace{5pt}
\noindent \textbf{Additional Analysis.}
We provide additional analysis in Appendix~\ref{appendix:additional_analysis}, covering verifier convergence, layer and steering-strength sensitivity, robustness to thought-type labeling, Pass@K sampling efficiency, and capacity-aware accuracy--efficiency analysis. For the latter, we report an Efficiency--Capacity (EC) score that combines relative accuracy gains, token reductions, and model size to characterize deployment trade-offs across scales. Overall, these analyses support the robustness of {\ourmethod}: the latent verifier converges quickly and remains stable under moderate perturbations, adaptive steering improves sampling efficiency, and EC trends align with the main results, with {\ourmethod} achieving the best overall trade-off.

\section{Conclusion}
We introduced {\ourmethod}, an adaptive latent steering framework for efficient LLM reasoning. {\ourmethod} uses a lightweight latent verifier to select among execution, reflection, transition, and no-intervention actions at thought boundaries. By distilling process-supervision signals into hidden-state quality estimates, the method avoids inference-time PRM calls and additional candidate decoding. Across mathematical reasoning and coding benchmarks, {\ourmethod} improves the accuracy--efficiency trade-off over vanilla decoding and fixed steering baselines. Our analyses further suggest that latent verification is a practical mechanism for controlling the reasoning behavior of LLMs at test time.

\section*{Limitations}
{\ourmethod} has several limitations. First, the intervention layer and steering strength are tuned on validation data rather than learned jointly with the verifier. Although mid-layer interventions with moderate strength perform well empirically, a more principled and adaptive calibration could improve robustness across models and tasks. Second, the method relies on a small discrete action space (execution, reflection, transition, no intervention). While this design ensures efficiency and interpretability, it restricts the expressivity of the steering policy compared to continuous or compositional alternatives. Third, the latent verifier is distilled from PRM supervision, making its reliability dependent on the quality and coverage of the underlying PRM. Although we observe transfer to out-of-domain math and coding benchmarks, generalization to broader settings such as planning, dialogue \cite{yan2025sharechat}, and long-form generation, remains underexplored. Finally, thought segmentation is based on paragraph boundaries and simple heuristics for constructing steering directions. This may be suboptimal; future work should consider learned segmentation and end-to-end optimization of intervention timing, action selection, and steering magnitude.

More broadly, {\ourmethod} may have implications for authorship privacy, as controllable latent interventions could modulate stylistic signals within loops of obfuscation, imitation, and verification an increasingly important concern in AI-assisted writing \cite{nguyen2025unraveling}.

\section*{Acknowledgments}
The authors acknowledge the use of ChatGPT and Grammarly for editorial assistance and ChatGPT for assistance with figure visualization.

\newpage
\clearpage

\bibliography{custom}

\begin{thebibliography}{45}
\providecommand{\natexlab}[1]{#1}

\bibitem[{Ahn et~al.(2024)Ahn, Verma, Lou, Liu, Zhang, and Yin}]{ahn2024large}
Janice Ahn, Rishu Verma, Renze Lou, Di~Liu, Rui Zhang, and Wenpeng Yin. 2024.
\newblock \href {https://aclanthology.org/2024.eacl-srw.17/} {Large language models for mathematical reasoning: Progresses and challenges}.
\newblock \emph{EACL}.

\bibitem[{AlphaProof and AlphaGeometry(2024)}]{alphaproof2024ai}
Team AlphaProof and Team AlphaGeometry. 2024.
\newblock \href {https://deepmind.google/blog/ai-solves-imo-problems-at-silver-medal-level} {Ai achieves silver-medal standard solving international 178 mathematical olympiad problems}.
\newblock \emph{DeepMind blog}.

\bibitem[{Azizi et~al.(2025)Azizi, Potraghloo, and Pedram}]{azizi2025activation}
Seyedarmin Azizi, Erfan~Baghaei Potraghloo, and Massoud Pedram. 2025.
\newblock \href {https://openreview.net/pdf?id=LLxSS9i2JD} {Activation steering for chain-of-thought compression}.
\newblock \emph{arXiv}.

\bibitem[{Besta et~al.(2024)Besta, Blach, Kubicek, Gerstenberger, Podstawski, Gianinazzi, Gajda, Lehmann, Niewiadomski, Nyczyk et~al.}]{besta2024graph}
Maciej Besta, Nils Blach, Ales Kubicek, Robert Gerstenberger, Michal Podstawski, Lukas Gianinazzi, Joanna Gajda, Tomasz Lehmann, Hubert Niewiadomski, Piotr Nyczyk, and 1 others. 2024.
\newblock \href {https://arxiv.org/pdf/2308.09687} {Graph of thoughts: Solving elaborate problems with large language models}.
\newblock In \emph{AAAI}.

\bibitem[{Brown et~al.(2024)Brown, Juravsky, Ehrlich, Clark, Le, R{\'e}, and Mirhoseini}]{brown2024large}
Bradley Brown, Jordan Juravsky, Ryan Ehrlich, Ronald Clark, Quoc~V Le, Christopher R{\'e}, and Azalia Mirhoseini. 2024.
\newblock \href {https://arxiv.org/pdf/2407.21787} {Large language monkeys: Scaling inference compute with repeated sampling}.
\newblock \emph{arXiv}.

\bibitem[{Burns et~al.(2023)Burns, Ye, Klein, and Steinhardt}]{burns2022discovering}
Collin Burns, Haotian Ye, Dan Klein, and Jacob Steinhardt. 2023.
\newblock \href {https://arxiv.org/pdf/2212.03827} {Discovering latent knowledge in language models without supervision}.
\newblock \emph{ICLR}.

\bibitem[{Chen et~al.(2025{\natexlab{a}})Chen, Zhang, Hong, Kundu, and Wang}]{chen2025seal}
Runjin Chen, Zhenyu Zhang, Junyuan Hong, Souvik Kundu, and Zhangyang Wang. 2025{\natexlab{a}}.
\newblock \href {https://arxiv.org/pdf/2504.07986} {Seal: Steerable reasoning calibration of large language models for free}.
\newblock \emph{COLM}.

\bibitem[{Chen et~al.(2025{\natexlab{b}})Chen, Zhao, Xia, Lu, Wang, Chen, Zhang, Wang, Li, and Shen}]{chen2025reasoning}
Xinghao Chen, Anhao Zhao, Heming Xia, Xuan Lu, Hanlin Wang, Yanjun Chen, Wei Zhang, Jian Wang, Wenjie Li, and Xiaoyu Shen. 2025{\natexlab{b}}.
\newblock \href {https://arxiv.org/abs/2505.16782} {Reasoning beyond language: A comprehensive survey on latent chain-of-thought reasoning}.
\newblock \emph{arXiv}.

\bibitem[{Chen et~al.(2025{\natexlab{c}})Chen, Xu, Liang, He, Pang, Yu, Song, Liu, Zhou, Zhang et~al.}]{chen2024not}
Xingyu Chen, Jiahao Xu, Tian Liang, Zhiwei He, Jianhui Pang, Dian Yu, Linfeng Song, Qiuzhi Liu, Mengfei Zhou, Zhuosheng Zhang, and 1 others. 2025{\natexlab{c}}.
\newblock \href {https://openreview.net/pdf?id=MSbU3L7V00} {Do not think that much for 2+ 3=? on the overthinking of o1-like llms}.
\newblock \emph{ICML}.

\bibitem[{Chen et~al.(2025{\natexlab{d}})Chen, Liu, Tian, Tong, Luo, and Liu}]{chen2025advancing}
Zui Chen, Tianqiao Liu, Mi~Tian, Qing Tong, Weiqi Luo, and Zitao Liu. 2025{\natexlab{d}}.
\newblock \href {https://arxiv.org/abs/2501.14002} {Advancing mathematical reasoning in language models: The impact of problem-solving data, data synthesis methods, and training stages}.
\newblock \emph{ICLR}.

\bibitem[{Cobbe et~al.(2021)Cobbe, Kosaraju, Bavarian, Chen, Jun, Kaiser, Plappert, Tworek, Hilton, Nakano et~al.}]{cobbe2021training}
Karl Cobbe, Vineet Kosaraju, Mohammad Bavarian, Mark Chen, Heewoo Jun, Lukasz Kaiser, Matthias Plappert, Jerry Tworek, Jacob Hilton, Reiichiro Nakano, and 1 others. 2021.
\newblock \href {https://arxiv.org/pdf/2110.14168} {Training verifiers to solve math word problems}.
\newblock \emph{arXiv}.

\bibitem[{Do et~al.(2025)Do, Hwang, Han, Oh, and Yun}]{do2025defines}
Heejin Do, Jaehui Hwang, Dongyoon Han, Seong~Joon Oh, and Sangdoo Yun. 2025.
\newblock \href {https://arxiv.org/pdf/2510.20603} {What defines good reasoning in llms? dissecting reasoning steps with multi-aspect evaluation}.
\newblock \emph{arXiv}.

\bibitem[{Fu et~al.(2025)Fu, Chen, Zhu, Fu, Dai, Zhuang, Ma, Qiao, Rosing, Stoica et~al.}]{fu2024efficiently}
Yichao Fu, Junda Chen, Siqi Zhu, Zheyu Fu, Zhongdongming Dai, Yonghao Zhuang, Yian Ma, Aurick Qiao, Tajana Rosing, Ion Stoica, and 1 others. 2025.
\newblock \href {https://openreview.net/pdf/4ca50f6ea693aeda7b95d22f1929d6a5d49cf4ff.pdf} {Efficiently scaling llm reasoning with certaindex}.
\newblock \emph{NeurIPS}.

\bibitem[{Fu et~al.(2026)Fu, Wang, Tian, and Zhao}]{fu2025deep}
Yichao Fu, Xuewei Wang, Yuandong Tian, and Jiawei Zhao. 2026.
\newblock \href {https://openreview.net/pdf?id=8LqHs0KIM7} {Deep think with confidence}.
\newblock \emph{ICLR}.

\bibitem[{Guo et~al.(2025)Guo, Yang, Zhang, Song, Zhang, Xu, Zhu, Ma, Wang, Bi et~al.}]{guo2025deepseek}
Daya Guo, Dejian Yang, Haowei Zhang, Junxiao Song, Ruoyu Zhang, Runxin Xu, Qihao Zhu, Shirong Ma, Peiyi Wang, Xiao Bi, and 1 others. 2025.
\newblock \href {https://arxiv.org/pdf/2501.12948} {Deepseek-r1: Incentivizing reasoning capability in llms via reinforcement learning}.
\newblock \emph{arXiv}.

\bibitem[{Hao et~al.(2025)Hao, Sukhbaatar, Su, Li, Hu, Weston, and Tian}]{hao2024training}
Shibo Hao, Sainbayar Sukhbaatar, DiJia Su, Xian Li, Zhiting Hu, Jason Weston, and Yuandong Tian. 2025.
\newblock \href {https://arxiv.org/pdf/2412.06769} {Training large language models to reason in a continuous latent space}.
\newblock \emph{COLM}.

\bibitem[{Hendrycks et~al.(2021)Hendrycks, Burns, Kadavath, Arora, Basart, Tang, Song, and Steinhardt}]{hendrycks2021measuring}
Dan Hendrycks, Collin Burns, Saurav Kadavath, Akul Arora, Steven Basart, Eric Tang, Dawn Song, and Jacob Steinhardt. 2021.
\newblock \href {https://openreview.net/forum?id=7Bywt2mQsCe} {Measuring mathematical problem solving with the math dataset}.
\newblock \emph{NeurIPS Datasets and Benchmarks Track}.

\bibitem[{Houlsby et~al.(2019)Houlsby, Giurgiu, Jastrzebski, Morrone, De~Laroussilhe, Gesmundo, Attariyan, and Gelly}]{houlsby2019parameter}
Neil Houlsby, Andrei Giurgiu, Stanislaw Jastrzebski, Bruna Morrone, Quentin De~Laroussilhe, Andrea Gesmundo, Mona Attariyan, and Sylvain Gelly. 2019.
\newblock Parameter-efficient transfer learning for nlp.
\newblock In \emph{International conference on machine learning}, pages 2790--2799. PMLR.

\bibitem[{Jain et~al.(2025)Jain, Gu, Li, Yan, Zhang, Wang, Solar-Lezama, Sen, and Stoica}]{jain2025livecodebench}
Naman Jain, Alex Gu, Wen-Ding Li, Fanjia Yan, Tianjun Zhang, Sida Wang, Armando Solar-Lezama, Koushik Sen, and Ion Stoica. 2025.
\newblock \href {https://openreview.net/pdf?id=chfJJYC3iL} {Livecodebench: Holistic and contamination free evaluation of large language models for code}.
\newblock In \emph{International Conference on Learning Representations}.

\bibitem[{Jin et~al.(2025)Jin, Yu, Huang, Zeng, Wang, Hua, Zhao, Mei, Meng, Ding et~al.}]{jin2025exploring}
Mingyu Jin, Qinkai Yu, Jingyuan Huang, Qingcheng Zeng, Zhenting Wang, Wenyue Hua, Haiyan Zhao, Kai Mei, Yanda Meng, Kaize Ding, and 1 others. 2025.
\newblock \href {https://aclanthology.org/2025.coling-main.37.pdf} {Exploring concept depth: How large language models acquire knowledge and concept at different layers?}
\newblock In \emph{COLING}.

\bibitem[{Lightman et~al.(2024)Lightman, Kosaraju, Burda, Edwards, Baker, Lee, Leike, Schulman, Sutskever, and Cobbe}]{lightman2023let}
Hunter Lightman, Vineet Kosaraju, Yuri Burda, Harrison Edwards, Bowen Baker, Teddy Lee, Jan Leike, John Schulman, Ilya Sutskever, and Karl Cobbe. 2024.
\newblock \href {https://openreview.net/pdf?id=v8L0pN6EOi} {Let's verify step by step}.
\newblock In \emph{ICLR}.

\bibitem[{Liu et~al.(2024)Liu, Kong, Liu, and Sun}]{liu2024fantastic}
Zhu Liu, Cunliang Kong, Ying Liu, and Maosong Sun. 2024.
\newblock \href {https://aclanthology.org/2024.findings-acl.866.pdf} {Fantastic semantics and where to find them: Investigating which layers of generative llms reflect lexical semantics}.
\newblock \emph{ACL Findings}.

\bibitem[{Malik et~al.(2026)Malik, Pyatkin, Land, Morrison, Smith, Hajishirzi, and Lambert}]{malik2025rewardbench}
Saumya Malik, Valentina Pyatkin, Sander Land, Jacob Morrison, Noah~A Smith, Hannaneh Hajishirzi, and Nathan Lambert. 2026.
\newblock \href {https://openreview.net/pdf?id=fb0G86Dewb} {Rewardbench 2: Advancing reward model evaluation}.
\newblock \emph{ICLR}.

\bibitem[{{Mathematical Association of America}(2023)}]{amc2023}
{Mathematical Association of America}. 2023.
\newblock \href {https://www.maa.org/math-competitions/amc-1012-contests} {American mathematics competitions (amc) 10 and 12, 2023}.
\newblock Problems and Answer Keys.

\bibitem[{Nguyen et~al.(2025)Nguyen, Hu, and Le}]{nguyen2025unraveling}
Tuc Nguyen, Yifan Hu, and Thai Le. 2025.
\newblock Unraveling interwoven roles of large language models in authorship privacy: Obfuscation, mimicking, and verification.
\newblock In \emph{EMNLP}.

\bibitem[{Nguyen and Le(2024{\natexlab{a}})}]{nguyen2024generalizability}
Tuc Nguyen and Thai Le. 2024{\natexlab{a}}.
\newblock Generalizability of mixture of domain-specific adapters from the lens of signed weight directions and its application to effective model pruning.
\newblock In \emph{ACL}.

\bibitem[{Nguyen and Le(2026)}]{nguyen2026beyond}
Tuc Nguyen and Thai Le. 2026.
\newblock Beyond linear activation steering: Invertible latent transformations for controlling llm behavior.
\newblock \emph{arXiv}.

\bibitem[{Nguyen and Le(2024{\natexlab{b}})}]{nguyen-le-2024-adapters}
Tuc~Van Nguyen and Thai Le. 2024{\natexlab{b}}.
\newblock \href {https://doi.org/10.18653/v1/2024.emnlp-main.1180} {Adapters mixup: Mixing parameter-efficient adapters to enhance the adversarial robustness of fine-tuned pre-trained text classifiers}.
\newblock In \emph{Proceedings of the 2024 Conference on Empirical Methods in Natural Language Processing}, Miami, Florida, USA. Association for Computational Linguistics.

\bibitem[{Stolfo et~al.(2025)Stolfo, Balachandran, Yousefi, Horvitz, and Nushi}]{stolfo2024improving}
Alessandro Stolfo, Vidhisha Balachandran, Safoora Yousefi, Eric Horvitz, and Besmira Nushi. 2025.
\newblock \href {https://openreview.net/pdf?id=wozhdnRCtw} {Improving instruction-following in language models through activation steering}.
\newblock \emph{ICLR}.

\bibitem[{Sun et~al.(2025)Sun, Min, Chen, Zhao, Fang, Liu, Wang, and Wen}]{sun2025challenging}
Haoxiang Sun, Yingqian Min, Zhipeng Chen, Wayne~Xin Zhao, Lei Fang, Zheng Liu, Zhongyuan Wang, and Ji-Rong Wen. 2025.
\newblock \href {https://arxiv.org/pdf/2503.21380} {Challenging the boundaries of reasoning: An olympiad-level math benchmark for large language models}.
\newblock \emph{arXiv}.

\bibitem[{Team(2024)}]{qwq-32b-preview}
Qwen Team. 2024.
\newblock \href {https://qwenlm.github.io/blog/qwq-32b-preview/} {Qwq: Reflect deeply on the boundaries of the unknown}.

\bibitem[{Turner et~al.(2023)Turner, Thiergart, Leech, Udell, Vazquez, Mini, and MacDiarmid}]{turner2023steering}
Alexander~Matt Turner, Lisa Thiergart, Gavin Leech, David Udell, Juan~J Vazquez, Ulisse Mini, and Monte MacDiarmid. 2023.
\newblock \href {https://arxiv.org/pdf/2308.10248} {Steering language models with activation engineering}.
\newblock \emph{arXiv}.

\bibitem[{Valmeekam et~al.(2023)Valmeekam, Marquez, Sreedharan, and Kambhampati}]{valmeekam2023planning}
Karthik Valmeekam, Matthew Marquez, Sarath Sreedharan, and Subbarao Kambhampati. 2023.
\newblock \href {https://openreview.net/pdf?id=X6dEqXIsEW} {On the planning abilities of large language models-a critical investigation}.
\newblock \emph{NeurIPS}.

\bibitem[{Wang et~al.(2024)Wang, Li, Shao, Xu, Dai, Li, Chen, Wu, and Sui}]{wang2024math}
Peiyi Wang, Lei Li, Zhihong Shao, Runxin Xu, Damai Dai, Yifei Li, Deli Chen, Yu~Wu, and Zhifang Sui. 2024.
\newblock \href {https://aclanthology.org/2024.acl-long.510.pdf} {Math-shepherd: Verify and reinforce llms step-by-step without human annotations}.
\newblock In \emph{ACL}.

\bibitem[{Wang et~al.(2025)Wang, Liu, Xu, Liang, Chen, He, Song, Yu, Li, Zhang et~al.}]{wang2025thoughts}
Yue Wang, Qiuzhi Liu, Jiahao Xu, Tian Liang, Xingyu Chen, Zhiwei He, Linfeng Song, Dian Yu, Juntao Li, Zhuosheng Zhang, and 1 others. 2025.
\newblock \href {https://arxiv.org/pdf/2501.18585} {Thoughts are all over the place: On the underthinking of o1-like llms}.
\newblock \emph{ICML}.

\bibitem[{Wei et~al.(2022)Wei, Wang, Schuurmans, Bosma, Xia, Chi, Le, Zhou et~al.}]{wei2022chain}
Jason Wei, Xuezhi Wang, Dale Schuurmans, Maarten Bosma, Fei Xia, Ed~Chi, Quoc~V Le, Denny Zhou, and 1 others. 2022.
\newblock \href {https://openreview.net/pdf?id=_VjQlMeSB_J} {Chain-of-thought prompting elicits reasoning in large language models}.
\newblock \emph{NeurIPS}.

\bibitem[{Xia et~al.(2025)Xia, Deng, Dunn, and Zhang}]{10.1145/3715754}
Chunqiu~Steven Xia, Yinlin Deng, Soren Dunn, and Lingming Zhang. 2025.
\newblock \href {https://arxiv.org/pdf/2407.01489} {Demystifying llm-based software engineering agents}.
\newblock \emph{The ACM on Software Engineering}.

\bibitem[{Yan et~al.(2025)Yan, Nguyen, Su, Lieffers, and Le}]{yan2025sharechat}
Yueru Yan, Tuc Nguyen, Bo~Su, Melissa Lieffers, and Thai Le. 2025.
\newblock Sharechat: A dataset of chatbot conversations in the wild.
\newblock \emph{arXiv}.

\bibitem[{Yang et~al.(2024)Yang, Yang, Hui, Zheng, Yu, Zhou, Li, Li, Liu, Huang, Dong, Wei, Lin, Tang, Wang, Yang, Tu, Zhang, Ma, Xu, Zhou, Bai, He, Lin, Dang, Lu, Chen, Yang, Li, Xue, Ni, Zhang, Wang, Peng, Men, Gao, Lin, Wang, Bai, Tan, Zhu, Li, Liu, Ge, Deng, Zhou, Ren, Zhang, Wei, Ren, Fan, Yao, Zhang, Wan, Chu, Liu, Cui, Zhang, and Fan}]{qwen2}
An~Yang, Baosong Yang, Binyuan Hui, Bo~Zheng, Bowen Yu, Chang Zhou, Chengpeng Li, Chengyuan Li, Dayiheng Liu, Fei Huang, Guanting Dong, Haoran Wei, Huan Lin, Jialong Tang, Jialin Wang, Jian Yang, Jianhong Tu, Jianwei Zhang, Jianxin Ma, and 40 others. 2024.
\newblock \href {https://arxiv.org/pdf/2407.10671} {Qwen2 technical report}.
\newblock \emph{arXiv}.

\bibitem[{Yao et~al.(2023)Yao, Yu, Zhao, Shafran, Griffiths, Cao, and Narasimhan}]{yao2023tree}
Shunyu Yao, Dian Yu, Jeffrey Zhao, Izhak Shafran, Tom Griffiths, Yuan Cao, and Karthik Narasimhan. 2023.
\newblock \href {https://openreview.net/pdf?id=5Xc1ecxO1h} {Tree of thoughts: Deliberate problem solving with large language models}.
\newblock \emph{NeurIPS}.

\bibitem[{Yoon et~al.(2025)Yoon, Kim, Yang, Kim, Kim, Kim, Choi, Kim, and Seo}]{yoon2025reasoning}
Dongkeun Yoon, Seungone Kim, Sohee Yang, Sunkyoung Kim, Soyeon Kim, Yongil Kim, Eunbi Choi, Yireun Kim, and Minjoon Seo. 2025.
\newblock \href {https://openreview.net/pdf?id=rbBtoVnduo} {Reasoning models better express their confidence}.
\newblock \emph{NeurIPS}.

\bibitem[{Yue et~al.(2025)Yue, Chen et~al.}]{yue2025does}
Yang Yue, Zhiqi Chen, and 1 others. 2025.
\newblock \href {https://openreview.net/pdf?id=4OsgYD7em5} {Does reinforcement learning really incentivize reasoning capacity in llms beyond the base model?}
\newblock \emph{arXiv}.

\bibitem[{Zhang et~al.(2025)Zhang, Wu, Zhou, Wu, Zhang, Ponnusamy, Subbaraj, Wang, Song, and Athiwaratkun}]{zhang2025understanding}
Zhenyu Zhang, Xiaoxia Wu, Zhongzhu Zhou, Qingyang Wu, Yineng Zhang, Pragaash Ponnusamy, Harikaran Subbaraj, Jue Wang, Shuaiwen~Leon Song, and Ben Athiwaratkun. 2025.
\newblock \href {https://arxiv.org/pdf/2512.24574} {Understanding and steering the cognitive behaviors of reasoning models at test-time}.
\newblock In \emph{NeurIPS Workshop on Efficient Reasoning}.

\bibitem[{Zhou et~al.(2023)Zhou, Sch{\"a}rli, Hou, Wei, Scales, Wang, Schuurmans, Cui, Bousquet, Le et~al.}]{zhou2022least}
Denny Zhou, Nathanael Sch{\"a}rli, Le~Hou, Jason Wei, Nathan Scales, Xuezhi Wang, Dale Schuurmans, Claire Cui, Olivier Bousquet, Quoc Le, and 1 others. 2023.
\newblock \href {https://openreview.net/pdf?id=WZH7099tgfM} {Least-to-most prompting enables complex reasoning in large language models}.
\newblock \emph{ICLR}.

\bibitem[{Zhu et~al.(2025)Zhu, Peng, Cheng, Qu, Huang, Zhu, Wang, Xue, Zhang, Shan, Cai, Kergan, Kembay, Smith, Lin, Nguyen, Pan, Chou, Cai, Wu, Zhao, Liu, Yang, Zhou, Zheng, Li, Zhou, Li, Zhang, Liu, Zhang, Huang, and Eshraghian}]{zhu2025surveylatentreasoning}
Rui-Jie Zhu, Tianhao Peng, Tianhao Cheng, Xingwei Qu, Jinfa Huang, Dawei Zhu, Hao Wang, Kaiwen Xue, Xuanliang Zhang, Yong Shan, Tianle Cai, Taylor Kergan, Assel Kembay, Andrew Smith, Chenghua Lin, Binh Nguyen, Yuqi Pan, Yuhong Chou, Zefan Cai, and 14 others. 2025.
\newblock \href {https://arxiv.org/pdf/2507.06203} {A survey on latent reasoning}.

\end{thebibliography}

\newpage
\clearpage
\appendix
\tableofcontents
\clearpage
\twocolumn

\section{Appendix}
\label{sec:appendix}

\subsection{Potential Risks}
\label{appendix:broader_impact}
{\ourmethod} provides a lightweight mechanism for controlling LLM reasoning behavior at test time without updating model parameters. By adaptively selecting latent steering actions, it can improve reasoning efficiency, reduce redundant generation, and lower inference cost while maintaining or improving final-answer accuracy.  These benefits may make reasoning models more practical in resource-constrained deployments. However, stronger test-time control mechanisms can also be misused.  The same steering actions that promote concise or corrective reasoning could be used to induce undesirable behaviors, bias model outputs, or weaken safety-oriented responses if optimized for harmful objectives. Deployment should therefore include safeguards such as access control for steering vectors and verifier checkpoints, monitoring of applied interventions, misuse-oriented evaluation, and transparent reporting when steering is used. We view {\ourmethod} as a reasoning-control framework that requires careful auditing and alignment between the verifier objective and the intended application.

\subsection{Full thought-label keyword list}
\label{thoughts_classification}
Table \ref{tab:criteria} shows how thoughts can be categorized into different groups.
\begin{table}[htb]
    \footnotesize
    \centering
    \renewcommand{\arraystretch}{1.2}
    \begin{tabular}{p{1.5cm} | p{5cm}}
        \toprule
        \textbf{Transition} 
        & ``alternatively'', ``think differently'', ``another way'', ``another approach'', ``another method'', ``another solution'', ``another strategy'', ``another technique'' \\  
        \midrule
        \textbf{Reflection} 
        & ``wait'', ``verify'', ``make sure'', ``hold on'', ``think again'', ``'s correct'', ``'s incorrect'', ``let me check'', ``seems right'' \\  
        \bottomrule
    \end{tabular}
    \caption{Criteria for recognizing transition and reflection thoughts}
    \label{tab:criteria}
\end{table}

\subsection{Repetitive Reasoning}
\label{repetitive_reasoning}
This section provides a qualitative example illustrating why fixed steering can be brittle in multi-step reasoning. In this case, applying a single execution-oriented steering mode encourages the model to continue direct computation, but the model fails to transition from intermediate arithmetic to the final averaging step. The actual generated output is shown in Figure~\ref{fig:repetitive_reasoning}, where the model repeatedly restates the same calculation for Beatrice and Marcell without progressing to the final answer. This behavior suggests that enforcing one reasoning mode throughout generation can over-constrain the reasoning trajectory and prevent the model from adapting when reflection or transition is needed. {\ourmethod} addresses this failure mode by selecting among execution, reflection, transition, and no-intervention actions at each thought boundary, allowing the model to adjust its reasoning behavior based on the current latent state.

\begin{figure}[tb]
\footnotesize
\centering
\begin{tcolorbox}[
  colback=gray!5,
  colframe=black,
  boxrule=0.5pt,
  arc=5pt,
  left=2pt,
  right=2pt,
  top=2pt,
  bottom=2pt
]
\textbf{Question:} Marcell and Beatrice are having a contest to see who can eat the most [...] Beatrice's was two roll-ups wide and 24 roll-ups long while Marcell's was 3 roll-ups witde and 14 roll-ups long. If they both ate their entire amount, how many did they eat on average? \\
\textbf{Output: } 
\textit{Okay, so Marcell and Beatrice are having a contest to see who can eat the most fruit roll-ups.} \\
.........\\
\textit{Wait, maybe the problem is saying that Beatrice's fruit roll-ups are two roll-ups wide and 24 roll-ups long, so each of her fruit roll-ups is $2 \times 24 = 48$. But Marcell's is $3 \times 14 = 42$.}
\textit{Wait, maybe the problem is saying that Beatrice's fruit roll-ups are two roll-ups wide and 24 roll-ups long, so each of her fruit roll-ups is $2 \times 24 = 48$. But Marcell's is $3 \times 14 = 42$.}
\textit{Wait, maybe the problem is saying that Beatrice's fruit roll-ups are two roll-ups wide and 24 roll-ups long, so each of her fruit roll-ups is $2 \times 24 = 48$. But Marcell's is $3 \times 14 = 42$.} \\
.........
\end{tcolorbox}
\caption{An example illustrating undesirable \textbf{repetitive reasoning} of Qwen-1.5B on GSM8K when applying a single execution-oriented steering mode, following the fixed steering setup of SEAL~\cite{chen2025seal}.}
\label{fig:repetitive_reasoning}
\vspace{-10pt}
\end{figure}

\subsection{Prompt Usage}
\label{prompt_usage}
All experiments use the same prompt template to elicit step-by-step reasoning:
\begin{tcolorbox}[
  colback=gray!5,
  colframe=black,
  boxrule=0.5pt,
  arc=5pt,
  left=2pt,
  right=2pt,
  top=2pt,
  bottom=2pt
]
\begin{verbatim}
Please reason step by step, and put your
final answer within \boxed{}.
User: {problem}
Assistant: <think>
\end{verbatim}
\end{tcolorbox}
We append the \texttt{<think>} token without a closing tag so that reasoning models generate an explicit intermediate reasoning trace before producing the final answer. This also provides a consistent structure for extracting hidden states at paragraph boundaries, marked by \texttt{\textbackslash n\textbackslash n}, which we use as thought-level intervention points throughout our experiments.

\subsection{Hyper-parameter Settings}
\paragraph{Steering Vector Extraction.}
We extract steering action vectors offline from 1{,}000 QA pairs sampled from
the GSM8K and MATH training splits. Following SEAL~\cite{chen2025seal}, we
segment each generated reasoning trace into thoughts using paragraph-boundary
tokens, marked by \texttt{\textbackslash n\textbackslash n}
(\texttt{ĊĊ} in the tokenizer vocabulary). At each boundary, we extract the
hidden representation from the intervention layer. Each thought is assigned a
weak thought-type label corresponding to execution, reflection, or transition.

Let \(\mathcal{I}_E\), \(\mathcal{I}_R\), and \(\mathcal{I}_T\) denote the
sets of thought boundaries labeled as execution, reflection, and transition,
respectively. For each mode \(c\in\{E,R,T\}\), we compute its mean hidden
representation:
\[
\bar{z}_c^{(\ell)}
=
\frac{1}{|\mathcal{I}_c|}
\sum_{(x,t)\in \mathcal{I}_c}
z_t^{(\ell)}(x).
\]
We then construct candidate action vectors by contrasting each target mode
against the complementary modes:
\[
\begin{aligned}
v_E^{(\ell)}
&=
\bar{z}_E^{(\ell)}
-
\bar{z}_{R\cup T}^{(\ell)},\\
v_R^{(\ell)}
&=
\bar{z}_R^{(\ell)}
-
\bar{z}_{E\cup T}^{(\ell)},\\
v_T^{(\ell)}
&=
\bar{z}_T^{(\ell)}
-
\bar{z}_{E\cup R}^{(\ell)},\\
v_{\varnothing}^{(\ell)}
&=
\mathbf{0}.
\end{aligned}
\]
Here, \(\bar{z}_{R\cup T}^{(\ell)}\), \(\bar{z}_{E\cup T}^{(\ell)}\), and
\(\bar{z}_{E\cup R}^{(\ell)}\) are the mean hidden representations over the
corresponding complementary sets of thought boundaries. Thus, each non-null
action vector encourages its target reasoning mode while suppressing the other
two modes. The execution action \(v_E^{(\ell)}\) reduces to the SEAL-style
execution steering vector, while \(v_R^{(\ell)}\) and \(v_T^{(\ell)}\) extend
the same contrastive mechanism to reflection and transition. The resulting
action vectors are saved in GGUF format and applied during inference using
additive \texttt{direct} steering without normalization.

\paragraph{Latent Verifier Architecture and Training.}
The latent verifier is a lightweight feedforward network that predicts a continuous step-quality score from hidden states at the steering layer. For Qwen-1.5B, it is a two-hidden-layer MLP with dimensions \(1536 \!\rightarrow\! 256 \!\rightarrow\! 128 \!\rightarrow\! 1\), ReLU activations, dropout rate 0.2, and a sigmoid output. The verifier has 426{,}497 parameters, corresponding to approximately 0.03\% of the 1.5B base model. It is trained to regress continuous step-level quality scores produced by Math-Shepherd, using a 60\%/20\%/20\% train/validation/test split.

\begin{table}[ht]
\centering
\footnotesize
\begin{tabular}{ll}
\toprule
\textbf{Hyper-parameter} & \textbf{Value} \\
\midrule
Epochs              & 100 \\
Learning rate       & $1 \times 10^{-4}$ \\
Optimizer           & Adam \\
Weight decay        & $1 \times 10^{-5}$ \\
Batch size          & 64 \\
Loss function       & MSE \\
Dropout             & 0.2 \\
Early stopping      & Validation loss \\
Gradient clipping   & Max norm 1.0 \\
\bottomrule
\end{tabular}
\caption{Latent verifier training hyper-parameters.}
\label{tab:verifier_hyperparams}
\end{table}

\paragraph{Adaptive Steering at Inference.}
During inference, the model periodically evaluates candidate steering configurations at thought boundaries and selects the action with the highest verifier score. We use greedy decoding and a maximum generation length of 8{,}192 tokens.

\begin{table}[ht]
\centering
\footnotesize
\begin{tabular}{ll}
\toprule
\textbf{Parameter} & \textbf{Value} \\
\midrule
Max tokens                    & 8192 \\
Temperature                   & 0 \\
Context safety margin          & 256 tokens \\
Maximum chunk size             & 512 tokens \\
Maximum transition/problem & 2 \\
Early-exit threshold           & 0.8 \\
Default initial mode           & \texttt{execution\_only} \\
\bottomrule
\end{tabular}
\caption{Adaptive inference hyper-parameters.}
\label{tab:adaptive_inference_hyperparams}
\end{table}
At each checkpoint, candidate actions are scored using algebraic latent
evaluation rather than generate-then-score verification. Specifically, for each candidate action \(a \in \{\varnothing,E,R,T\}\), the verifier scores the perturbed hidden state
\[
\hat{\mathbf{h}}_a =
\mathbf{h}_{\text{current}} + \alpha v_a^{(\ell)},
\]
where \(v_a^{(\ell)}\) is the action vector for action \(a\), and
\(v_{\varnothing}^{(\ell)}=\mathbf{0}\). This avoids generating
additional candidate continuations and makes action selection substantially
cheaper than text-level verification.

\paragraph{Robustness Evaluation.}
We evaluate verifier stability under hidden-state perturbations by adding isotropic Gaussian noise:
\[
\tilde{\mathbf{h}} = \mathbf{h} + \boldsymbol{\epsilon}, 
\qquad
\boldsymbol{\epsilon} \sim \mathcal{N}(\mathbf{0}, \sigma^2 \mathbf{I}).
\]
Noise magnitudes are scaled by the empirical hidden-state variance:
\[
\sigma = \lambda \cdot \overline{\operatorname{std}}(\mathbf{h}),
\]
where \(\overline{\operatorname{std}}(\mathbf{h})\) is the mean standard deviation across the 1{,}536 hidden dimensions and \(\lambda \in \{0.0, 0.05, 0.10, 0.25, 0.50, 1.00, 2.00\}\). Each noise level is evaluated over 10 independent runs on 500 held-out reasoning steps.

\subsection{Benchmark, Model, and Evaluation Details}
\label{appendix:experimental_details}

\paragraph{Benchmarks.}
We evaluate {\ourmethod} on five mathematical reasoning benchmarks and one coding benchmark. 
GSM8K~\citep{cobbe2021training} contains grade-school math word problems that require multi-step arithmetic reasoning. 
MATH~\citep{hendrycks2021measuring} is the full 5000-problem test set covering competition-level topics, including algebra, geometry, number theory, counting and probability, and calculus. 
AMC2023~\citep{amc2023}, AIME2024, and AIME2025~\citep{sun2025challenging} provide contest-style problems that require more advanced mathematical reasoning and often involve longer solution chains. 
To evaluate transfer beyond mathematics, we also include LiveCodeBench~\citep{jain2025livecodebench}, a temporally held-out code-generation benchmark designed to reduce benchmark contamination. Detailed statistics are provided in Table \ref{tab:datasets}.

\begin{table}[t]
\centering
\footnotesize
\begin{tabular}{cc}
\toprule
\textbf{Benchmark} 
& \textbf{Size}   \\
\midrule
GSM8K  & 1319  \\
MATH   & 5000 \\
AMC2023 &  83  \\
AIME2024 & 30  \\
AIME2025  & 30 \\
LiveCodeBench & 400 \\
\bottomrule
\end{tabular}
\vspace{-2pt}
\caption{Evaluation benchmarks used in our experiments. The suite covers grade-school arithmetic, competition-level mathematics, olympiad-style contest problems, and code generation. Difficulty labels indicate the intended problem level and are used only as descriptive metadata.}
\label{tab:datasets}
\vspace{-5pt}
\end{table}

\paragraph{Models.}
We evaluate several widely used reasoning models: DeepSeek-R1-Distill-Qwen-1.5B, 7B, and 32B, denoted as R1-Distill-1.5B/7B/32B~\citep{guo2025deepseek}, as well as QwQ-32B-Preview~\citep{qwq-32b-preview,qwen2}. 
For all models and methods, we use the same instruction prompt and set the maximum generation length to 8{,}192 tokens. 
This ensures that performance differences are attributable to the steering policy rather than differences in prompting or decoding budget.

\paragraph{Metrics.}
We report final-answer accuracy (\textit{Acc.}), defined as the percentage of examples for which the extracted final answer matches the ground-truth answer. 
For mathematical benchmarks, we use answer extraction based on the final boxed or explicitly stated answer. 
For LiveCodeBench, correctness is determined by the benchmark's standard execution-based evaluation. 
We also report the average number of generated tokens (\textit{\#Tok}) as a measure of inference efficiency, excluding prompt tokens. 
A method is considered to achieve a better accuracy--efficiency trade-off when it improves final-answer accuracy while reducing, or at least not substantially increasing, generated token usage.

\subsection{Detailed Baselines}
\label{appendix:detailed_baselines}
We compare {\ourmethod} against inference-time baselines covering no intervention, fixed activation steering, and head-level steering. All methods use the same prompt template and evaluation benchmarks described in Section~\ref{prompt_usage}. Unless otherwise specified, decoding is greedy with a maximum length of 8{,}192 tokens.

\paragraph{Vanilla.}
Vanilla uses the base reasoning model without any hidden-state, attention-head, or logit-level intervention. This setting provides the reference point for all reported changes in accuracy and token usage. We evaluate Vanilla across all target models, including DeepSeek-R1-Distill-Qwen-1.5B, 7B, 32B, and QwQ-32B-Preview.

\paragraph{SEAL~\citep{chen2025seal}.}
SEAL is a static activation-steering baseline that encourages execution-oriented reasoning throughout generation. Following the execution/reflection/transition decomposition, we apply the fixed \texttt{execution\_only} configuration at paragraph boundaries, marked by \texttt{\textbackslash n\textbackslash n}. Specifically, SEAL applies the fixed execution action vector
\(v_E^{(\ell)}=\bar{z}_E^{(\ell)}-\bar{z}_{R\cup T}^{(\ell)}\) throughout
generation, without adaptive action selection. The steering vectors are extracted offline from 1{,}000 training traces. SEAL is the closest fixed-policy counterpart to {\ourmethod}; it tests whether a single execution-promoting direction is sufficient, without adaptive action selection.

\paragraph{ASC~\citep{azizi2025activation}.}
ASC is a fixed activation-steering baseline designed to reduce inefficient or overly verbose reasoning. It applies a precomputed steering direction during decoding without conditioning on the quality of intermediate reasoning states. We include ASC to compare {\ourmethod} with a static steering approach that targets reasoning efficiency but does not select among multiple reasoning modes.

\paragraph{CREST~\citep{zhang2025understanding}.}
CREST performs head-level steering by identifying attention heads associated with inefficient cognitive behaviors and suppressing the corresponding directions at inference time. Unlike SEAL and {\ourmethod}, which intervene on hidden states using reasoning-mode vectors, CREST operates at the attention-head level and applies a static suppression policy. This baseline evaluates whether adaptive hidden-state steering provides benefits beyond static multi-head intervention.

\paragraph{ATLAS(T).}
ATLAS(T) is a text-verifier variant of our method. At each thought boundary, it selects among candidate steering actions using a text-level verifier rather than the latent verifier used by {\ourmethod}. This variant provides a direct comparison between explicit text-level verification and latent verification. Because ATLAS(T) requires scoring generated reasoning text, it is substantially more expensive than latent scoring, but it serves as a useful reference for evaluating whether hidden-state verification can approximate text-based adaptive control.

\subsection{Qualitative Analysis: Correlation with Downstream Accuracy}
\label{appendix:qualitative_correlation}
We provide a qualitative example illustrating how verifier-guided adaptive steering correlates with improved reasoning outcomes in Figure \ref{fig:verifier_correlation_example}. In this case, fixed steering leads to repetitive reasoning without progress, while ATLAS dynamically adjusts the reasoning mode and produces a correct solution. This example highlights that the verifier can direct the model generation toward more coherent reasoning trajectories that progress toward correct solutions. In contrast, low-quality reasoning steps (e.g., repetitive execution without transition) receive lower scores and are corrected through adaptive steering. This supports the use of the latent verifier as a proxy signal for guiding reasoning decisions at test time.

\begin{figure*}[tb]
\footnotesize
\centering
\begin{tcolorbox}[
  colback=gray!5,
  colframe=black,
  boxrule=0.5pt,
  arc=5pt,
  left=2pt,
  right=2pt,
  top=2pt,
  bottom=2pt
]
\textbf{Question:} Marcell and Beatrice are having a contest to see who can eat the most fruit roll-ups. Beatrice's was two roll-ups wide and 24 roll-ups long, while Marcell's was 3 roll-ups wide and 14 roll-ups long. If they both ate their entire amount, how many did they eat on average?
\end{tcolorbox}

\vspace{2pt}
\begin{tcolorbox}[
  colback=red!5,
  colframe=red!60!black,
  boxrule=0.5pt,
  arc=5pt,
  left=2pt,
  right=2pt,
  top=2pt,
  bottom=2pt
]
\textbf{Failure:} \\
\textit{Okay, so Marcell and Beatrice are having a contest to see who can eat the most fruit roll-ups.} \\
\textit{Wait, maybe the problem is saying that Beatrice's fruit roll-ups are $2 \times 24 = 48$, but Marcell's is $3 \times 14 = 42$.} \\
\textit{Wait, maybe the problem is saying that Beatrice's fruit roll-ups are $2 \times 24 = 48$, but Marcell's is $3 \times 14 = 42$.} \\
\textit{Wait, maybe the problem is saying that Beatrice's fruit roll-ups are $2 \times 24 = 48$, but Marcell's is $3 \times 14 = 42$.} \\
\textit{... (repeats without progressing to final answer)}
\end{tcolorbox}

\vspace{2pt}
\begin{tcolorbox}[
  colback=green!5,
  colframe=green!60!black,
  boxrule=0.5pt,
  arc=5pt,
  left=2pt,
  right=2pt,
  top=2pt,
  bottom=2pt
]
\textbf{ATLAS:} \\
Okay, so Marcell and Beatrice are having a contest to see who can eat the most fruit roll-ups. 
They unroll as many as they can find. But then someone made a mistake, and Beatrice's was two roll-ups wide and 24 rolls up long. 
Marcell's was 3 roll-ups wide and 14 roll-ups long. If they both ate their entire amount, how many did they eat on average? Hmm, let me try to figure this out. So, I think the key here is to find out how many fruit roll-ups each person ate and then find the average. 
Since they ate their entire amount, I guess it's just the total number each ate divided by two. First, let's tackle Beatrice's roll-ups. She has a roll that's two roll-ups wide and 24 rolls long. 
So, I think that means it's a rectangle, right? So, the area would be width multiplied by length. So, 2 times 24. Let me write that down: 2 * 24. That should give me the total number of roll-ups she ate. Calculating that, 2 times 24 is 48. 
So, Beatrice ate 48 roll-ups. Okay, that seems straightforward.Now, moving on to Marcell's roll-ups. He has a roll that's 3 roll-ups wide and 14 rolls long. Again, that should be a rectangle, so the area is width times length. So, 3 times 14. Let me compute that: 3 * 14. Hmm, 3 times 10 is 30, and 3 times 4 is 12, so 30 + 12 is 42. So, Marcell ate 42 roll-ups.Wait, let me double-check that. 3 times 14: 14 + 14 is 28, plus another 14 is 42. Yeah, that's correct. So, Marcell ate 42 roll-ups. Now, to find the average number they ate, I need to add up both totals and divide by 2. 
So, Beatrice had 48, Marcell had 42. Let me add those together: 48 + 42. That's 90. So, together, they ate 90 roll-ups. To find the average, I divide 90 by 2. 90 divided by 2 is 45. So, the average number of roll-ups they ate is 45. Wait, let me make sure I didn't make any mistakes. So, Beatrice: 2 * 24 is 48. Marcell: 3 * 14 is 42. Total: 48 + 42 is 90. Average: 90 / 2 is 45. 
Yeah, that seems right. I don't think I made any calculation errors here. It's pretty straightforward. So, the average number of roll-ups they ate is 45. **Final Answer** The average number of fruit roll-ups they ate is 45. Beatrice's roll-ups are 2 roll-ups wide and 24 rolls long, so the total number of roll-ups she ate is calculated as: (2 times 24 = 48) Marcell's roll-ups are 3 roll-ups wide and 14 rolls long, so the total number of roll-ups he ate is calculated as: 3 times 14 = 42. To find the average number of roll-ups they ate, we add the totals together and divide by 2: {Total} = 48 + 42 = 90. {Average} = frac\{90\}\{2\} = 45. The average number of fruit roll-ups they ate is 45.
\end{tcolorbox}

\caption{Example illustrating how adaptive steering corrects low-quality reasoning. Fixed steering leads to repetitive execution without progress, while ATLAS dynamically adjusts reasoning behavior and produces a correct solution. This demonstrates that higher verifier scores correlate with more coherent and effective reasoning trajectories.}
\label{fig:verifier_correlation_example}
\vspace{-10pt}
\end{figure*}

\subsection{Detailed Results for Text-Verifier ATLAS}
\label{appendix:atlas_t_results}
Tables~\ref{tab:atlas_t_indomain_appendix} and~\ref{tab:atlas_t_crossdomain_appendix} report the detailed performance of ATLAS(T), the text-verifier variant of our method. Table~\ref{tab:atlas_t_indomain_appendix} shows results on in-domain benchmarks, while Table~\ref{tab:atlas_t_crossdomain_appendix} reports cross-domain results. We include final-answer accuracy and average generated tokens for each model--benchmark pair to provide a complete view of the accuracy--efficiency trade-off achieved by text-level verification.

\begin{table}[t]
\centering
\footnotesize
\setlength{\tabcolsep}{5pt}
\begin{tabular}{lcccc}
\toprule
\multirow{2}{*}{Model}
& \multicolumn{2}{c}{MATH}
& \multicolumn{2}{c}{GSM8K} \\
\cmidrule(lr){2-3} \cmidrule(lr){4-5}
& Acc. & \#Tok & Acc. & \#Tok \\
\midrule
R1-Distill-1.5B & 82.20 & 2783 & 85.29 & 1452 \\
R1-Distill-7B   & 90.70 & 2975 & 89.46 & 672  \\
R1-Distill-32B  & 93.02 & 1937 & 93.48 & 473  \\
QwQ-32B-Pre     & 90.40 & 1962 & 95.07 & 511  \\
\bottomrule
\end{tabular}
\caption{In-domain performance of ATLAS(T).}
\label{tab:atlas_t_indomain_appendix}
\end{table}

\begin{table*}[t]
\centering
\footnotesize
\begin{tabular}{lcccccccccc}
\toprule
\multirow{2}{*}{Model}
& \multicolumn{2}{c}{GSM8K}
& \multicolumn{2}{c}{AIME2024}
& \multicolumn{2}{c}{AMC2023}
& \multicolumn{2}{c}{AIME2025}
& \multicolumn{2}{c}{LiveCodeBench} \\
\cmidrule(lr){2-3} \cmidrule(lr){4-5} \cmidrule(lr){6-7}
\cmidrule(lr){8-9} \cmidrule(lr){10-11}
& Acc. & \#Tok
& Acc. & \#Tok
& Acc. & \#Tok
& Acc. & \#Tok
& Acc. & \#Tok \\
\midrule
R1-Distill-1.5B & 84.46 & 1189 & 30.00 & 6361 & 62.50 & 3843 & 26.67 & 6279 & 31.00 & 5890 \\

R1-Distill-7B & 89.46 & 623 & 60.00 & 5721 & 82.50 & 3752 & 40.00 & 6245
& 38.50 & 5850 \\

R1-Distill-32B & 93.93 & 419 & 66.67 & 5438 & 87.50 & 3084 & 43.33 & 5816
& 49.00 & 4450 \\

QwQ-32B-Pre & 95.22 & 438 & 60.00 & 5532 & 87.50 & 3482 & 43.33 & 5832 & 51.00 & 4200 \\
\bottomrule
\end{tabular}%

\caption{Cross-domain performance of ATLAS(T) with text-level verification.}
\label{tab:atlas_t_crossdomain_appendix}
\end{table*}


\subsection{Latent Verifier Network Training Cost}
\label{appendix:verifier_training_cost}
Table~\ref{tab:verifier_training_cost} reports the architecture size and training time of the latent verifier for each target model. 
Across all settings, we use the same two-hidden-layer StepLevelMLP:
\[
\begin{aligned}
&\mathrm{FC}(d \rightarrow 256) \rightarrow \mathrm{ReLU} \rightarrow \mathrm{Dropout}(0.2) \\
&\rightarrow \mathrm{FC}(256 \rightarrow 128) \rightarrow \mathrm{ReLU} \rightarrow \mathrm{Dropout}(0.2) \\
&\rightarrow \mathrm{FC}(128 \rightarrow 1) \rightarrow \mathrm{Sigmoid},
\end{aligned}
\]
where \(d\) denotes the hidden dimension of the target model at the intervention layer. 
Thus, only the input projection varies across base models, while the remaining verifier architecture is shared. 
The verifier is trained to regress PRM-provided step-quality scores using MSE loss, Adam optimizer with learning rate \(1\times10^{-4}\), weight decay \(1\times10^{-5}\), batch size 64, dropout 0.2, and gradient clipping with maximum norm 1.0.  Although we train for up to 100 epochs, validation loss typically plateaus around epoch 20, making the effective training cost even smaller in practice. 
Overall, the verifier adds negligible offline overhead: even for 32B-scale target models, it contains only 1.34M parameters and takes approximately one minute to train on a single H100 GPU.

\begin{table}[t]
\centering
\footnotesize
\setlength{\tabcolsep}{2.5pt}
\renewcommand{\arraystretch}{1.12}
\begin{tabular}{lccc}
\toprule
\textbf{Base Model}
& \textbf{Hidden Dim}
& \textbf{Params}
& \textbf{Train Time} \\
\midrule
R1-Distill-Qwen-1.5B & 1536  & 426K  & 43.5 second \\
R1-Distill-Qwen-7B & 3584 & 951K &  61.7 second \\
R1-Distill-Qwen-32B & 5120 & 1.34M & 62.9 second \\
QwQ-32B-Preview & 5120 & 1.34M & 64.3 second \\
\bottomrule
\end{tabular}
\caption{Latent verifier architecture size and training time across target models. All verifiers use the same StepLevelMLP architecture; only the input dimension changes with the hidden size of the base model. Training time reports 100-epoch wall-clock time on a single H100 GPU with batch size 64.}
\label{tab:verifier_training_cost}
\end{table}

\subsection{Computational Resources}
\label{computational_resource}
All experiments were conducted using CUDA 12.8. 
Experiments with R1-Distill-Qwen-1.5B were run on a single NVIDIA L40S GPU, while experiments with larger models, including R1-Distill-Qwen-7B, R1-Distill-Qwen-32B, and QwQ-32B-Preview, were conducted on a single NVIDIA H100 GPU.

\section{Additional Analysis}
\label{appendix:additional_analysis}
\subsection{Verifier Convergence}
\label{verifier_convergence}
\begin{figure}[tb]
    \centering
    \includegraphics[width=0.4\textwidth]{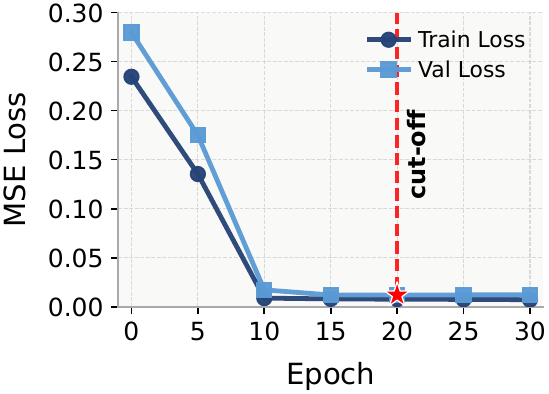}
    \caption{Training convergence of the latent verifier. Training and validation MSE decrease rapidly and stabilize around epoch 20, which we use as the cutoff point for verifier selection.}
    \label{fig:verifier_convergence}
\end{figure}

Figure~\ref{fig:verifier_convergence} shows the training dynamics of the latent verifier. Both training and validation losses drop sharply within the first 10 epochs and stabilize around epoch 20, where the validation MSE reaches its minimum value of 0.0119. This rapid convergence suggests that hidden states contain sufficient information to approximate PRM-provided step-level quality scores with a lightweight MLP. The validation loss remains stable after the cutoff point, indicating limited overfitting and supporting the use of the verifier for downstream adaptive steering.

\subsection{Stability Under Perturbation}
We evaluate whether the latent verifier remains reliable under small shifts in hidden-state representations, as adaptive steering scores candidate perturbed states during inference. Since our steering vectors are computed from mean differences between naturally occurring reasoning modes, the resulting interventions are expected to stay within a locally meaningful representation subspace rather than introducing arbitrary off-manifold noise. To test this empirically, we add Gaussian perturbations of varying magnitudes to held-out hidden states and measure the absolute change in verifier prediction, \(|\Delta \mathrm{Score}|\). The perturbation variance is estimated from the hidden-state distribution, with \(\sigma^2 = 3.07\). As shown in Table~\ref{tab:verifier_stability}, the verifier is stable under moderate perturbations: prediction changes remain below 0.01 for noise scales up to 0.25 and remain small even under larger perturbations. These results suggest that the verifier provides a robust ranking signal for candidate steering actions during adaptive inference. We provide a qualitative example of adaptive steering correcting low-quality reasoning in Appendix~\ref{appendix:qualitative_correlation}.

\begin{table}[t]
\centering
\footnotesize
\setlength{\tabcolsep}{1pt}
\renewcommand{\arraystretch}{0.95}
\begin{tabular}{lccccccc}
\toprule
Scale & 0.00 & 0.05 & 0.10 & 0.25 & 0.50 & 1.00 & 2.00 \\
\midrule
$|\Delta$Score$|$ 
& 0.0000 & 0.0014 & 0.0030 & 0.0070 & 0.0296 & 0.0279 & 0.0566 \\
\bottomrule
\end{tabular}
\vspace{-4pt}
\caption{Verifier stability under Gaussian perturbations.}
\label{tab:verifier_stability}
\vspace{-8pt}
\end{table}

\subsection{Layer Ablation Study.}
For the layer ablation study, we randomly sample 1,000 examples from the MATH dataset. Prior work suggests that shallow layers primarily encode low-level lexical features. In contrast, middle layers capture higher-level semantic and conceptual representations that are more closely associated with reasoning processes \cite{liu2024fantastic, jin2025exploring, chen2025seal}.  Motivated by these findings, we focus our ablation on the middle layers of each model to determine where steering interventions have the most impact. Specifically, for R1-Distill-1.5B and R1-Distill-7B, which each contain 28 layers, we evaluate interventions applied to layers in the range $[15, 25]$. For larger models, including R1-Distill-32B and QwQ-32B-Preview, we perform ablations over layers $[45, 60]$. For each configuration, we record both task accuracy and test-time token usage to analyze the trade-offs between effectiveness and efficiency. 
\label{layer_ablation_study}
\begin{figure}[tb]
    \centering
    \includegraphics[width=0.495\textwidth]{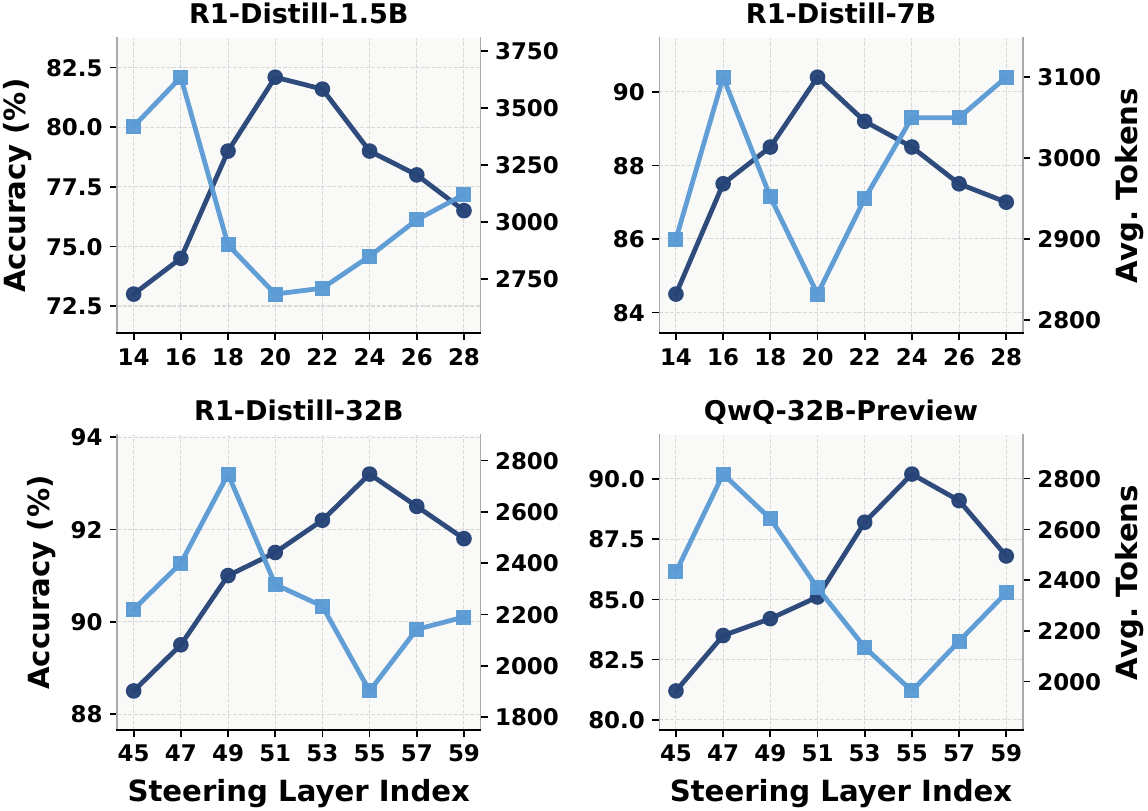}
    \caption{Layer-wise ablation study of steering effectiveness. \textcolor{darkblue}{Dark blue} indicates the accuracy achieved by {\ourmethod}, while \textcolor{lightskyblue}{sky blue} denotes the corresponding test-time token usage. }
    \label{fig:layer_ablation_study}
\end{figure}
Figure~\ref{fig:layer_ablation_study} shown a layer-wise ablation to identify which transformer layers are most effective for intervention-based steering. Overall, interventions applied to middle layers consistently yield the most effective reasoning behavior, achieving higher accuracy while reducing test-time token usage. This observation aligns with prior findings that middle layers primarily encode abstract and conceptual knowledge in large language models \cite{jin2025exploring}. These results highlight layer selection as a critical factor in adaptive steering and motivate the incorporation of layer-aware strategies into the proposed dynamic steering mechanism.

\subsection{Strength ablation.}
We vary the steering strength \(\alpha\) to study the sensitivity of ATLAS to intervention magnitude. As shown in Table~\ref{tab:strength_ablation}, weak interventions provide limited control over the reasoning trajectory, while moderate strengths yield the best accuracy--efficiency trade-off. Performance generally peaks around \(\alpha=1.0\), with \(\alpha=1.25\) sometimes producing slightly shorter generations but comparable or marginally lower accuracy. Larger strengths degrade both accuracy and token efficiency, suggesting that overly strong interventions may push hidden states away from the verifier-calibrated region. We therefore use \(\alpha=1.0\) as the default setting in all main experiments.
\begin{table*}[t]
\centering
\footnotesize
\setlength{\tabcolsep}{2.6pt}
\begin{tabular}{lcccccccccccc}
\toprule
\multirow{2}{*}{\(\alpha\)}
& \multicolumn{2}{c}{\textbf{MATH}}
& \multicolumn{2}{c}{\textbf{GSM8K}}
& \multicolumn{2}{c}{\textbf{AIME2024}}
& \multicolumn{2}{c}{\textbf{AMC2023}}
& \multicolumn{2}{c}{\textbf{AIME2025}}
& \multicolumn{2}{c}{\textbf{LiveCodeBench}} \\
\cmidrule(lr){2-3} \cmidrule(lr){4-5}
\cmidrule(lr){6-7} \cmidrule(lr){8-9}
\cmidrule(lr){10-11} \cmidrule(lr){12-13}
& Acc. $\uparrow$ & Tok. $\downarrow$
& Acc. $\uparrow$ & Tok. $\downarrow$
& Acc. $\uparrow$ & Tok. $\downarrow$
& Acc. $\uparrow$ & Tok. $\downarrow$
& Acc. $\uparrow$ & Tok. $\downarrow$
& Acc. $\uparrow$ & Tok. $\downarrow$ \\
\midrule
0.00 & 73.76 & 3941 & 79.30 & 2390 & 20.00 & 7396 & 47.50 & 5241 & 10.00 & 7893 & 19.00 & 8268 \\
0.25 & 76.90 & 3590 & 81.10 & 1970 & 23.33 & 7150 & 52.50 & 4900 & 13.33 & 7550
& 22.00 & 7800 \\
0.50 & 79.70 & 3210 & 82.75 & 1540 & 26.67 & 6880 & 57.50 & 4420 & 16.67 & 7050
& 26.50 & 7150 \\
0.75 & 81.45 & 2920 & 83.90 & 1295 & 30.00 & 6520 & 62.50 & 4020 & 20.00 & 6600
& 30.00 & 6600 \\
1.00 & \textbf{82.28} & 2754 & \textbf{84.53} & 1171 & \textbf{33.33} & 6304 & \textbf{65.00} & 3837 & \textbf{23.33} & 6337 & \textbf{32.00} & 6216 \\
1.25 & 82.05 & \textbf{2690} & 84.20 & \textbf{1135} & \textbf{33.33} & \textbf{6180} & 62.50 & \textbf{3750} & \textbf{23.33} & \textbf{6250} & 31.00 & \textbf{6120} \\
1.50 & 80.70 & 2830 & 83.10 & 1240 & 30.00 & 6450 & 60.00 & 3970 & 20.00 & 6500
& 28.50 & 6450 \\
2.00 & 78.00 & 3350 & 80.80 & 1680 & 23.33 & 7020 & 52.50 & 4680 & 16.67 & 7100
& 23.50 & 7400 \\
\bottomrule
\end{tabular}%
\caption{Steering-strength ablation on R1-Distill-Qwen-1.5B. We report final-answer accuracy and average generated tokens across in-domain and cross-domain benchmarks. \(\alpha=0\) corresponds to Vanilla decoding, while \(\alpha=1.0\) is the default setting used in the main experiments. Best values in each benchmark are bolded.}
\label{tab:strength_ablation}
\end{table*}

\subsection{Robustness of Thought-Type Labeling}
\label{sec:label_robustness}
Our approach relies on heuristic segmentation and keyword-based rules to assign reasoning steps into execution, reflection, and transition categories. While this provides a scalable form of weak supervision, such heuristics may introduce noise and raise concerns about robustness across models and prompts. To evaluate the reliability of this labeling strategy, we conduct a robustness check using a LLM (GPT-OSS-120B \footnote{https://huggingface.co/openai/gpt-oss-120b}) to independently classify thought types on reasoning traces generated by R1-Distill-1.5B. We observe an agreement rate of \textbf{88.32\%} between the heuristic labels and the LLM-based classifier, indicating strong alignment with a high-capacity model's judgment. We further assess the impact of labeling noise on downstream performance. Specifically, we re-extract hidden representations using the LLM-derived labels, retrain the latent verifier on GSM8K, and evaluate ATLAS on the test set. Results are shown in Table~\ref{tab:label_robustness}. The performance gap between heuristic and LLM-based labeling is minimal (85.37 vs. 84.91), suggesting that ATLAS is robust to reasonable variations in thought-type annotation. These findings indicate that simple heuristic labeling provides a reliable and scalable supervision signal, and that the proposed framework does not critically depend on precise annotation of thought types.
\begin{table}[t]
\centering
\footnotesize
\setlength{\tabcolsep}{6pt}
\begin{tabular}{lc}
\toprule
\textbf{Method} & \textbf{Accuracy} \\
\midrule
Baseline & 79.30 \\
ATLAS (heuristic labels) & \textbf{85.37} \\
ATLAS (LLM labels) & 84.91 \\
\bottomrule
\end{tabular}
\caption{Robustness of ATLAS to thought-type labeling strategies.}
\label{tab:label_robustness}
\vspace{-10pt}
\end{table}

\subsection{Accuracy-Efficiency Tradeoff} 
\label{accuracy_efficiency_tradeoff}
Beyond reporting standard accuracy metrics and test-time token usage as often done in existing work\cite{chen2025seal,azizi2025activation,zhang2025understanding}, we want to contextualize these metrics to meaningfully compare their final performance.
For instance, a method that achieves comparable accuracy while reducing token usage by 20\% on a larger model should be considered superior to alternatives that require substantially more computation. Motivated by this observation, we propose an \emph{Efficiency–Capacity Trade-off (EC) metric}, scaled from 0 to 100, that jointly accounts for accuracy, test-time token usage, and model size. This metric enables a more principled and deployment-aware comparison of steering methods under realistic computational constraints.
Specifically, given the relative change in performance (\(\Delta\)Acc) and token reduction (\(\Delta\)Token), we define the trade-off score of a method evaluated on model \(m\) as:
\begin{equation}
\text{EC}(m){=}\frac{1}{2}(\overline{\Delta \text{Acc}}(m){+}\overline{\Delta \text{Token}}(m){\times} \frac{P_m}{P_{\max}}),
\label{eq:ec_metric}
\end{equation}
where \(\overline{\Delta \text{Acc}}\), \(\overline{\Delta \text{Token}}\) denote the min--max normalized changes in accuracy and token usage, respectively, and \(P_m\), \(P_{\max}\) denote the \# parameters of model \(m\) and the largest model considered. 

The EC metric captures three key desiderata: (1) accuracy improvements are rewarded linearly through \(\overline{\Delta \text{Acc}}\); (2) token reductions are weighted by model capacity via the scaling factor \(\frac{P_m}{P_{\max}}\), reflecting the higher inference cost of larger models; and (3) min--max normalization ensures that both components contribute comparably despite differing scales. Higher EC scores indicate more favorable trade-offs between accuracy and efficiency. We report the \emph{EC score} in Table \ref{ec_score}. Overall, {\ourmethod} variants consistently achieve the highest \emph{EC} scores across all configurations, with \emph{ATLAS(L)} ranking first in 8 out of 12 cases and \emph{ATLAS(T)} leading in the remaining 4 on larger models. Notably, while CREST achieves superior accuracy on R1-Distill-7B (Table \ref{tab:crossdomain_performance}), its substantially higher token consumption results in poor efficiency-accuracy trade-offs relative to {\ourmethod} methods.

\begin{table*}[tb]
\centering
\footnotesize
\setlength{\tabcolsep}{3pt}
\begin{tabular}{lcccccccccccc}
\toprule
& \multicolumn{3}{c}{\text{R1-Distill-1.5B}} 
& \multicolumn{3}{c}{\text{R1-Distill-7B}}
& \multicolumn{3}{c}{\text{R1-Distill-32B}}
& \multicolumn{3}{c}{\text{QwQ-32B-Pre}} \\
\cmidrule(lr){2-4} \cmidrule(lr){5-7} \cmidrule(lr){8-10} \cmidrule(lr){11-13}
Methods & GSM8K & AIME & AMC & GSM8K & AIME & AMC & GSM8K & AIME & AMC & GSM8K & AIME & AMC \\
\midrule
\textit{Vanilla} 
& 22.4 & 22.1 & 11.5 
& 33.4 & 9.4 & 38.7 
& 74.0 & 5.5 & 17.0 
& 70.5 & 5.5 & 26.5 \\
\textit{SEAL}
& 45.4 & 30.0 & 22.7 
& 10.4 & 29.6 & 45.7 
& 56.0 & 39.0 & \underline{84.5} 
& 34.0 & 72.0 & 62.5 \\
\textit{ASC}
& 11.5 & 0.0 & 0.0 
& 11.8 & 3.9 & 1.3 
& 47.5 & 9.0 & 0.0 
& 50.5 & 7.5 & 0.0 \\
\textit{CREST}
& 1.4 & 29.6 & 33.9 
& \underline{50.0} & 7.0 & 7.5 
& 30.5 & 40.0 & 60.0 
& 46.0 & 44.0 & 56.0 \\
\textit{ATLAS(T)}
& \underline{51.8} & \underline{45.3} & \underline{46.9} 
& 46.4 & \textbf{60.9} & \underline{46.5} 
& \textbf{100.0} & \textbf{100.0} & \textbf{92.0} 
& \textbf{100.0} & \textbf{93.0} & \textbf{85.0} \\
\textit{ATLAS(L)}
& \textbf{52.4} & \textbf{52.4} & \textbf{52.4} 
& \textbf{50.5} & \underline{54.0} & \textbf{61.0} 
& \underline{93.0} & \underline{88.5} & 73.0 
& \underline{95.0} & \underline{79.5} & \underline{68.5} \\
\bottomrule
\end{tabular}
 \caption{EC scores across models and benchmarks. \textbf{Bold}, \underline{underlining} denote the best and second-best results.}
\label{ec_score}
\vspace{-8pt}
\end{table*}

\subsection{Sampling Efficiency}
\begin{figure}[tb]
    \centering
    \includegraphics[width=0.40\textwidth]{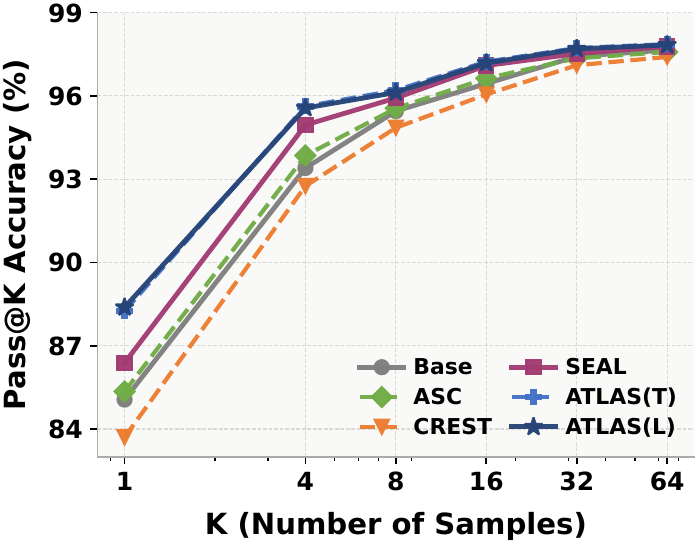}
    \caption{Pass@K performance of adaptive steering methods on the GSM8K dataset using R1-Distill-1.5B.}
    \label{fig:passk}
    \vspace{-5pt}
\end{figure}

We further evaluate sampling efficiency using \(\mathrm{Pass@}K\), which measures whether a model produces at least one correct solution among \(K\) sampled attempts~\cite{brown2024large}. Following recent reasoning evaluation protocols~\cite{yue2025does}, we report \(\mathrm{Pass@}K\) on GSM8K with R1-Distill-Qwen-1.5B. As shown in Figure~\ref{fig:passk}, {\ourmethod} consistently outperforms both vanilla decoding and static SEAL across all values of \(K\). The gains are most pronounced in the low-sample regime: at \(K=1\), {\ourmethod} achieves \(88.38\%\), compared with \(86.38\%\) for SEAL and \(85.06\%\) for the base model. As \(K\) increases, all methods approach saturation, but {\ourmethod} still maintains the highest \(\mathrm{Pass@}64\) score (\(97.83\%\)). These results suggest that adaptive latent steering not only improves single-sample accuracy but also shifts the sampling distribution toward more useful reasoning trajectories.

\subsection{Detail set ablation study}
\label{appendix_action_set_study}
Table~\ref{tab:action_set_ablation} reports the detailed action-set ablation results on R1-Distill-Qwen-1.5B. The execution-only setting $\{E\}$, which corresponds to the fixed SEAL-style policy, already improves over the no-intervention baseline, increasing accuracy from 73.76\% to 79.78\% on MATH and from 79.30\% to 82.41\% on GSM8K while substantially reducing token usage. However, adding adaptive choices consistently leads to stronger accuracy--efficiency trade-offs. In particular, incorporating the no-intervention action $\varnothing$ improves over fixed execution steering, suggesting that abstaining from unnecessary perturbations is important for avoiding over-steering. Adding reflection and transition actions further improves performance, showing that corrective and trajectory-shifting behaviors provide complementary benefits beyond execution-oriented progress. The full action set $\{\varnothing,E,R,T\}$ achieves the highest accuracy on both MATH and GSM8K, demonstrating that ATLAS benefits from selecting among multiple reasoning modes rather than relying on a single fixed steering direction.

\begin{table}[t]
\centering
\footnotesize
\setlength{\tabcolsep}{3.5pt}
\renewcommand{\arraystretch}{1.05}
\begin{tabular}{lcccc}
\toprule
\multirow{2}{*}{Action Set}
& \multicolumn{2}{c}{\textbf{MATH}}
& \multicolumn{2}{c}{\textbf{GSM8K}} \\
\cmidrule(lr){2-3} \cmidrule(lr){4-5}
& Acc. $\uparrow$ & Tok. $\downarrow$
& Acc. $\uparrow$ & Tok. $\downarrow$ \\
\midrule
\(\{\varnothing\}\) & 73.76 & 3941 & 79.30 & 2390 \\
\(\{E\}\) & 79.78 & 3034 & 82.41 & 1438 \\
\(\{R\}\) & 75.20 & 3850 & 79.53 & 2260 \\
\(\{T\}\) & 76.60 & 3605 & 80.44 & 2050 \\
\(\{\varnothing,E\}\)  & 80.64 & 2940 & 83.05 & 1365 \\
\(\{\varnothing,E,R\}\)  & 81.47 & 2865 & 83.78 & 1288 \\
\(\{\varnothing,E,T\}\) & 81.12 & 2810 & 83.61 & 1246 \\
\(\{E,R,T\}\) & \underline{81.76} & \underline{2795} & \underline{84.02} & \textbf{1208} \\
\(\{\varnothing,E,R,T\}\) & \textbf{82.28} & \textbf{2754} & \textbf{85.37} & \underline{1316} \\
\bottomrule
\end{tabular}
\vspace{-3pt}
\caption{Action-set ablation on R1-Distill-Qwen-1.5B. \(E\), \(R\), and \(T\) denote execution-, reflection-, and transition-oriented action vectors constructed by contrasting each target mode with its complement; \(\varnothing\) denotes no intervention. The execution-only setting \(\{E\}\) corresponds to the SEAL baseline. The best and second best results are \textbf{bolded} and \underline{underline}, respectively.}
\label{tab:action_set_ablation}
\vspace{-6pt}
\end{table}

\end{document}